\documentclass[letterpaper]{article} 
\usepackage{aaai2026}  
\usepackage{times}  
\usepackage{helvet}  
\usepackage{courier}  
\usepackage[hyphens]{url}  
\usepackage{graphicx} 
\usepackage{natbib}  
\usepackage{caption} 
\usepackage{algorithm}
\usepackage{stmaryrd}

\usepackage{newfloat}
\usepackage{listings}
\DeclareCaptionStyle{ruled}{labelfont=normalfont,labelsep=colon,strut=off} 
\floatstyle{ruled}
\newfloat{listing}{tb}{lst}{}
\floatname{listing}{Listing}
\nocopyright

\usepackage{xspace}

\usepackage{enumitem}

\usepackage{subcaption}

\usepackage{booktabs}       
\usepackage{amsfonts}       
\usepackage{nicefrac}       
\usepackage{microtype}      

\usepackage{multirow}

\usepackage{pifont}

\newcommand{\cmark}{\ding{51}}

\newcommand{\xmark}{\ding{55}}

\usepackage{algpseudocode}
\usepackage{amsmath}

\newcommand{\omicron}{o}
\usepackage{amsmath,amssymb}

\newcommand{\head}[1]{{\noindent\textbf{#1}}}

\newcommand{\tool}{\textsc{SmartyPat}\xspace}
\newcommand{\bench}{\textsc{SmartyPat-Bench}\xspace}
\newcommand{\soundbench}{\textsc{Benign-Bench}\xspace}
\newcommand{\augbench}{\textsc{SmartyPat-Bench-Augmented}\xspace}
\newcommand{\ruozhiba}{\textsc{COIG-CQIA}\xspace}

\usepackage{tabularx}
\newcolumntype{Y}{>{\centering\arraybackslash}X}
\usepackage{underscore}

\title{Socrates or Smartypants: Testing Logic Reasoning Capabilities of Large Language Models with Logic Programming-Based Test Oracles}

\author{
    Zihao Xu\textsuperscript{\rm 1}\equalcontrib,
    Junchen Ding\textsuperscript{\rm 1}\footnotemark[1],
    Yiling Lou\textsuperscript{\rm 2},
    Kun Zhang\textsuperscript{\rm 3},
    Dong Gong\textsuperscript{\rm 1},
    Yuekang Li\textsuperscript{\rm 1}\thanks{Corresponding author}
}
\affiliations{
    \textsuperscript{\rm 1}University of New South Wales, Australia, \{zhihao.xu2, junchen.ding, dong.gong, yuekang.li\}@unsw.edu.au\\
    \textsuperscript{\rm 2}Fudan University, China, yilinglou@fudan.edu.cn\\
    \textsuperscript{\rm 3}Carnegie Mellon University, USA, kunz1@cmu.edu\\

}

\usepackage{colortbl}

\begin{document}
\maketitle

\begin{abstract}
Large Language Models (LLMs) have achieved significant progress in language understanding and reasoning. Evaluating and analyzing their logical reasoning abilities has therefore become essential. However, existing datasets and benchmarks are often limited to overly simplistic, unnatural, or contextually constrained examples. 
In response to the growing demand, we introduce \bench{}, a challenging, naturally expressed, and systematically labeled benchmark derived from real-world high-quality Reddit posts containing subtle logical fallacies. Unlike existing datasets and benchmarks, it provides more detailed annotations of logical fallacies and features more diverse data. 
To further scale up the study and address the limitations of manual data collection and labeling, such as fallacy-type imbalance and labor-intensive annotation, we introduce \tool{}, an automated framework powered by logic programming-based oracles. 
\tool{} utilizes Prolog rules to systematically generate logically fallacious statements, which are then refined into fluent natural-language sentences by LLMs, ensuring precise fallacy representation. 
Extensive evaluation demonstrates that \tool{} produces fallacies comparable in subtlety and quality to human-generated content and significantly outperforms baseline methods. 
Finally, experiments reveal insights into LLM capabilities, highlighting that while excessive reasoning steps hinder fallacy detection accuracy, structured reasoning enhances fallacy categorization performance.
\end{abstract}

\section{Introduction}
\label{sec:intro}

LLMs demonstrate strong performance across diverse domains. As adoption increases, thorough evaluation across reasoning, domain knowledge, and problem-solving becomes crucial. Among these, logical reasoning is foundational, especially for tasks requiring structured thinking, such as programming. Prior studies explored LLMs' logical reasoning via symbolic-to-natural language conversion~\cite{han2022folio,parmar2024logicbench}, but these often yield rigid, unnatural text using formal constructs ($\forall, \exists$) rarely seen in human language. To address this, \citet{jin-etal-2022-logical} proposed the \textsc{LOGIC} dataset, short, realistic, but still simplistic fallacious statements from student quizzes labeled by fallacy type. Yet, many examples remain trivial. To address these limitations, researchers introduced the COIG-CQIA benchmark~\cite{coig_cqia}, which features subtle logical errors drawn from forum posts on the Chinese platform \textit{ruozhiba}. To adapt this content to English, \citet{zhai2025ruozhibench} proposed a translated benchmark. However, its reliance on direct Chinese-to-English translation weakens context-sensitive nuances critical for evaluating logical reasoning. Moreover, the lack of annotated fallacy types limits the applicability for robust fallacy categorization.

A key challenge remains: building a benchmark that is challenging, derived from real-world English data, and annotated with specific fallacy types. We introduce \bench{}, derived from a Reddit community~\cite{reddit_shittyaskscience} analogous to the Chinese forum ruozhiba. We manually reviewed 2,500 posts (by upvotes), removing low-quality content and selecting 502 posts for annotation. Our dataset highlights two key desiderata: \ding{182} Controllable generation of subtle logical fallacies. Manual datasets are imbalanced, three fallacy types dominate 79.7\%, while three rare ones comprise just 1.77\%. Generation offers better balance and quality. \ding{183} Fidelity to intended fallacy types. Accurate generation eases labeling and improves benchmark reliability.
To address these issues, we propose \tool{}, an automated generator of logically fallacious statements for LLM evaluation. \tool{} derives Prolog rules and fact structures from \bench{}, capturing the core logic of each fallacy type. LLMs then generate diverse fact instances conforming to these structures, which are verified and composed into natural language via Prolog, combining symbolic rigor with neural diversity and fluency, yielding a synthetic dataset, \augbench{}.

\tool{} produces high-quality, subtly fallacious statements comparable to human-written posts, outperforming two baselines: direct LLM generation and LLM-generated Prolog. We assess LLMs on fallacy detection and categorization. In the detection task, reasoning models tend to underperform due to overanalysis, leading to high FPR and lower F1 scores, while they generally excel at categorization.

In summary, we make the following contributions:
\begin{enumerate}[topsep=0pt, partopsep=0pt, itemsep=0pt, parsep=0pt]
\item We present \bench{}, the first benchmark comprising 502 high-quality, real-world English statements labeled with fine-grained logical fallacy types.
\item We introduce \tool{}, the first Prolog-based neural symbolic generation framework that synthesizes logically fallacious statements with built-in test oracles for evaluating LLM reasoning.
\item We conduct the first comprehensive evaluation of nine state-of-the-art LLMs on logical fallacy detection and classification, uncovering critical reasoning and alignment limitations.
\end{enumerate}

We release our code and data at GitHub\footnote{\url{https://github.com/ltroin/Smarty}}, and additionally provide an extended version \cite{xu2025socrates}. All appendices mentioned below refer to this extended version.

\section{Related work}
\head{Testing Deep Learning Models.}
DeepGauge~\cite{deepgauge} pioneered testing deep learning (DL) systems, emphasizing the importance of test oracles for DL, including LLMs. Subsequent works adapted traditional software testing techniques, such as mutation testing~\cite{Humbatova2021DeepCrimeMT} and fuzzing~\cite{Xie2019DeepHunterAC}, to DL systems. Recently, logic programming was introduced for generating logically sound factual knowledge to test LLMs for hallucination detection~\cite{drowzee}, highlighting its potential for effective LLM testing.

\head{Logical Fallacy Categorization}
Currently, there is no universally agreed-upon classification scheme for logical fallacies, and existing categorizations often include overlapping concepts. 
For instance, \citet{li2025llms} introduce the category \textit{Lame Jokes}, representing failures to grasp general knowledge or common sense, alongside \textit{Factual Error}, which similarly involves misunderstandings of basic facts. 
These two categories substantially overlap, potentially leading to inconsistencies in annotation. 
Similarly, \citet{zhai2025ruozhibench} define \textit{Logical Error} as contradictions or flawed reasoning, \textit{Commonsense Misunderstanding} as mistakes about everyday facts, and \textit{Erroneous Assumption} as incorrect premises. 
However, all these definitions could reasonably fall under a broader category such as \textit{Logical Error}, introducing ambiguity for both LLM interpretation and human annotation.

By analyzing existing classifications alongside the fallacious statements in our \bench{} dataset, we propose a refined categorization of logical fallacies with 14 distinct types, as shown in Table \ref{fig:fallacy_dist}. Further details on this categorization are provided in the appendix.

\label{sec:fallacy_cate}
\section{A Preliminary Study}
\label{sec:preli}
\label{sec:logicform}
\label{sec:datacharacter}

\begin{table}[!htb]
\centering
\small
\begin{tabular}{lccccc}
\toprule
\textbf{Benchmark}                                          & \textbf{NE}   & \textbf{CQ} & \textbf{RW}       & \textbf{FL}    \\ \midrule
FOLIO~\cite{han2022folio}                  & \cmark & \xmark     & \xmark     & \xmark     \\
P-FOLIO~\cite{han2024p}            & \cmark & \xmark     & \xmark     & \xmark     \\
LogicBench~\cite{parmar2024logicbench}     & \cmark & \xmark     & \xmark     & \xmark     \\
ContextHub~\cite{hua2024disentangling}    & \cmark & \xmark     & \xmark     & \xmark     \\
LOGIC~\cite{jin2022logical}               & \cmark & \xmark     & \xmark     & \cmark     \\ 
LFUD~\cite{li2024reason}                 & \cmark & \xmark     & \xmark     & \cmark     \\ 
BIG-Bench~\cite{srivastava2023beyond}    & \cmark & \xmark     & \xmark     & \cmark     \\
LogicAsker~\cite{wan2024logicasker}        & \cmark & \xmark     & \xmark     & \cmark     \\ 
\ruozhiba~\cite{coig_cqia}                & \xmark & \cmark     & \cmark     & \xmark     \\ 
RuoZhiBench~\cite{zhai2025ruozhibench}               & \xmark & \cmark     & \cmark     & \cmark     \\
\bench (This work)                         & \cmark & \cmark     & \cmark     & \cmark     \\
\bottomrule
\end{tabular}
\caption{Comparison of Benchmarks. The abbreviations denote: NE(Native English), CQ(Cunning Question), RW(Real World), FL(Fallacy Label)}
\label{tab:bench_comparsion}
\end{table}

\paragraph{Limitations of Existing Benchmarks}
Researchers have extensively explored the logical reasoning capabilities of LLMs, often through benchmarks grounded in symbolic logic. Datasets such as \textsc{Folio}, \textsc{P-FOLIO} (an enhanced version of \textsc{Folio} featuring artificially constructed sentences), \textsc{LogicBench}, and \textsc{ContextHub} employ synthetic constructions, e.g., using logical operators like \(\wedge\) and \(\vee\) to connect propositions, in order to evaluate deductive reasoning. While syntactically rigorous, these formats deviate from natural language and fail to reflect the kinds of reasoning errors encountered in real-world discourse. For example, applying Disjunctive Syllogism in such settings primarily assesses rule-based symbolic manipulation. In contrast, detecting a \textit{False Cause} fallacy, as studied in this work, requires semantic and commonsense reasoning, such as identifying when a correlation is incorrectly interpreted as causation.

\textsc{LogicAsker} extends the symbolic paradigm with annotations but still lacks coverage of semantically rich fallacies like \textit{False Cause}. More naturalistic efforts such as \textsc{LOGIC} attempt to bridge the gap by sourcing fallacy examples from student exam quizzes. However, these instances are often trivial, overly simplistic, and disconnected from real-world reasoning, thus providing limited challenge to LLMs. Other benchmarks, like \textsc{LFUD}, generate fallacies by prompting LLMs with isolated propositions (e.g., \textit{``Peter visited China last year''}) sourced from Wikipedia or textbooks. Yet these are typically not genuine fallacies and lack coherence or context. Moreover, the fallacy types used in \textsc{LFUD} diverge from those found in natural discourse. Our experiments confirm that generating realistic fallacies poses a much greater challenge for LLMs, as it demands deeper semantic and commonsense reasoning.

The COIG-CQIA benchmark improves realism by sourcing subtle, context-dependent fallacies from Chinese online forums. However, its reliance on direct Chinese--English translations~\cite{zhai2025ruozhibench} reduces contextual fidelity, and the lack of explicit fallacy labels limits its diagnostic value. Additionally, its fallacy categorization scheme suffers from conceptual ambiguity, with several categories lacking clear definitions or logical coherence (see Section~\ref{sec:fallacy_cate}).

Table~\ref{tab:bench_comparsion} presents a comparative overview of existing logic reasoning benchmarks for LLM evaluation. The limitations discussed above motivate the design of our benchmark.

\paragraph{Construction of \bench{}}
To construct \bench{}, we curated logically flawed posts from an English-language subreddit~\cite{reddit_shittyaskscience} analogous to \ruozhiba{}. Using the Arctic Shift dataset~\cite{heitmann_arctic_shift}, we extracted 251,052 entries and applied keyword filtering, upvote-based selection, and expert annotation, resulting in 502 high-quality examples. Each post was labeled with relevant fallacy types and manually transformed into a logic-constrained form $(p_1 \land \dots \land p_n) \rightarrow q$ for analysis. Further details are provided in the technical report.

\begin{table}[!hbtp]
    \centering
    \small
    \begin{tabular}{@{} l l r r @{}}
        \toprule
        \textbf{Abb.} & \textbf{Fallacy Type} & \textbf{Count} & \textbf{Percent(\%)} \\
        \midrule
        FP & False Premise                          & 218 & 35.12 \\
        EC & Equivocation                           & 189 & 30.40 \\
        FA & False Analogy                          & 88  & 14.17 \\
        NF & Nominal Fallacy                        & 38  & 6.12  \\
        CT & Contextomy                             & 32  & 5.16  \\
        FS & False Cause                            & 11  & 1.77  \\
        AF & Accident Fallacy                       & 8   & 1.29  \\
        ID & Improper Distribution or Addition      & 7   & 1.13  \\
        BQ & Begging the Question                   & 7   & 1.13  \\
        IE & Inverse Error                          & 6   & 0.97  \\
        WD & Wrong Direction                        & 6   & 0.97  \\
        FD & False Dilemma                          & 5   & 0.81  \\
        FC & Fallacy of Composition                 & 3   & 0.48  \\
        IT & Improper Transposition                 & 3   & 0.48  \\
        \bottomrule
    \end{tabular}
    \caption{Fallacy Types and Their Frequencies. The results indicate that the dataset exhibits a substantial class imbalance.}
    \label{fig:fallacy_dist}
\end{table}

\paragraph{Observations from \bench{}}

Table~\ref{fig:fallacy_dist} illustrates the distribution of various fallacy types within \bench{}. We observe that \textit{FP}, \textit{EC}, and \textit{FA} are the three most prevalent fallacies, collectively accounting for over 79.7\% of the dataset. 
In contrast, the three least frequent types, \textit{IT}, \textit{FC}, and \textit{FD}, together constitute only 1.77\%. 
This indicates a highly imbalanced representation of fallacy types in user-generated forum posts.
Additionally, the entire process of constructing a rigorous benchmark for evaluating logic reasoning from real-world data is labor-intensive and time-consuming. 
On average, screening each sentence required approximately 0.5 minutes, annotating the logical fallacy types took roughly 3 minutes, and transforming questions into declarative sentences took around 2 minutes. 
This resulted in a cumulative workload of approximately 3,760 minutes, or roughly 62.67 hours, for a single annotator.
Consequently, \textbf{to reduce manual effort in developing larger datasets and to maintain complete control over dataset content, there is a need for techniques capable of automatically generating high-quality, logically fallacious statements}.

\section{Methodology}
\label{sec:method}

The overall workflow of \textbf{\tool{}} is formalized in Algorithm~\ref{alg:unified} and illustrated in the appendix. The method comprises three stages: \ding{182}\textbf{PrologProgramDesign} (Section~\ref{sec:prologdesign}): Based on the logical implication format defined in Section~\ref{sec:logicform}, we analyze the structural properties of each fallacy type to derive schematic patterns, guiding the design of corresponding \textit{Prolog} predicates. This results in a complete Prolog program capable of supporting systematic knowledge generation (Lines 5–7). \ding{183}\textbf{PrologKnowledgeGeneration} (Section~\ref{sec:knowledgeneration}): Using fallacy-specific predicates from the previous stage, this module invokes an LLM to generate \textit{facts}, which are then injected into the Prolog knowledge base (Lines 8–9). \ding{184}\textbf{FallacySentenceTransformation} (Section~\ref{sec:sentencetransfomration}): The enriched Prolog knowledge base is executed to infer fallacy-specific outputs. Only fact combinations satisfying the logical criteria of the target fallacy are retained. These outputs are then transformed into natural language sentences and evaluated to ensure alignment with the intended fallacy type (Lines 10–13).

\begin{algorithm}[!t]
\small
\caption{\tool}
\label{alg:unified}
\begin{algorithmic}[1]
\Require $T_{decl}$: DeclarativeTemplates, $\Phi$: FallacyTypes, $\mathcal{L}$: LLM, $\Pi$: PrologEngine
\Ensure $S_{fallacy}$: GeneratedFallaciousSentences

\Function{\tool}{$T_{decl}, \Phi, \mathcal{L}, \Pi$}
    \State $\mathcal{K}_{prolog} \gets \emptyset$
    \Comment{Initialize Prolog knowledge base}

    \State $S_{fallacy} \gets \emptyset$
    \Comment{Initialize fallacious sentence set}

    \For{$\phi \in \Phi$}
        \State $T_{\phi} \gets$ \Call{FilterByFallacyType}{$T_{decl}, \phi$}
        \Comment{Filter relevant templates by type $\phi$}

        \State $\phi_{syn} \gets$ \Call{AnalyzeAndConstructSchema}{$T_{\phi}$}
        \Comment{Construct synthesized schema}

        \State $(\tilde{F}_{\phi}, \tilde{R}_{\phi}) \gets$ \Call{ExtractFactsRules}{$\phi_{syn}$}
        \Comment{Extract initial facts and rules from schema}

        \State $\tilde{F}_{add} \gets$ \Call{GenerateFactsWithLLM}{$\tilde{F}_{\phi}, \tilde{R}_{\phi}, \mathcal{L}$}
        \Comment{Expand facts using LLM}

        \State $\mathcal{K}_{prolog} \gets \mathcal{K}_{prolog} \cup \tilde{F}_{add} \cup \tilde{R}_{\phi}$

        \State $\Pi$.assertz($\mathcal{K}_{prolog}$)
        \Comment{Load all facts and rules into Prolog engine}

        \State $\mathcal{H}_{\tilde{F}_{\text{valid}}} \gets \textit{findall}(\llbracket \tilde{R}_{\phi} \rrbracket_{\mathcal{K}_{prolog}})$
        \Comment{Query Prolog to retrieve valid fact combinations}


        \State $s_{nl} \gets$ \Call{GenerateNLWithLLM}{$\mathcal{H}_{\tilde{F}_{\text{valid}}}, T_{phi}, \mathcal{L}$}
        \Comment{Generate NL sentences from valid facts}

        \State $S_{fallacy} \gets S_{fallacy} \cup \{s_{nl}\}$
    \EndFor

    \State \Return $S_{fallacy}$
\EndFunction

\end{algorithmic}
\end{algorithm}

\begin{table}[!ht]
\centering
\small
\begin{tabular}{|p{0.3\columnwidth}p{0.43\columnwidth}p{0.125\columnwidth}|}
\hline
\textbf{Predicate} & \textbf{Description} & \textbf{Notation} \\
\hline
has_effect($\alpha$, $\delta$, $\epsilon$) & An action $\alpha$ lasting $\delta$ results in $\epsilon$. & HE/3 \\
\hline
valid_accumulate($\upsilon$, $\rho$, $\tau$) & Repeating an action of duration $\upsilon$, satisfying $\rho$, accumulates to valid result of $\tau$. & VC/3 \\
\hline
established_fact($\chi$, $\phi$) & Condition $\chi$ establishes fact $\phi$. & EF/2 \\
\hline
false_premise($\phi$, $\pi$) & Fact $\phi$ has a false premise $\pi$. & FP/2 \\
\hline
plausible_observation($\omicron$, $\pi$) & Valid observation $\omicron$ can incorrectly justify $\pi$. & PO/2 \\
\hline
false_premise_lead_conclusion($\pi$, $\omicron$, $\gamma$) & False premise $\pi$ along with valid observation $\omicron$ leads to erroneous conclusion $\gamma$. & FPLC/3 \\
\hline
has_rule($\omicron$, $\rho$) & Object $\omicron$ contains instruction or rule $\rho$. & HR/2 \\
\hline
rule_unreasonable_interpretation($\rho$, $\iota$) & Rule $\rho$ could be unreasonably interpreted as $\iota$. & RUI/2 \\
\hline
rule_reasonable_interpretation($\rho$, $\iota$) & Rule $\rho$ could be reasonably interpreted as $\iota$. & RRI/2 \\
\hline
has_property($\chi$, $\pi$) & Component $\chi$ has property $\pi$. & HP/2 \\
\hline
is_part_of($\chi$, $\omega$) & Component $\chi$ is part of whole $\omega$. & IPO/2 \\
\hline
lacks_property($\omega$, $\pi$) & Whole $\omega$ lacks property $\pi$. & LP/2 \\
\hline
claim_and_argument($\chi$, $\alpha$) & Argument $\alpha$ is used to support claim $\chi$. & CA/2 \\
\hline
explicit_meaning_of_argument($\alpha$, $\epsilon$) & Argument $\alpha$ explicitly means $\epsilon$. & EMA/2 \\
\hline
explicit_meaning_rely_on_claim($\epsilon$, $\chi$) & Explicit meaning $\epsilon$ relies on claim $\chi$. & EMRC/2 \\
\hline
quote_context($\theta$, $\mu$) & Original meaning $\mu$ of quote $\theta$. & QC/2 \\
\hline
quote_out_of_context($\theta$, $\mu$) & Quote $\theta$ is misinterpreted as $\mu$. & QOC/2 \\
\hline
fact_related_out_of_context($\mu$, $\phi$) & Misinterpretation $\mu$ is related improperly to fact $\phi$. & FROC/2 \\
\hline
improper_fact_quote_out_of_context($\phi$, $\gamma$) & Fact $\phi$ improperly leads to conclusion $\gamma$. & IFQOC/2 \\
\hline
complement_cases($\alpha$, $\beta$) & Cases $\alpha$ and $\beta$ complement each other. & CC/2 \\
\hline
implies($\chi$, $\rho$) & Condition $\chi$ logically implies $\rho$. & IM/2 \\
\hline
cause($\alpha$, $\beta$) & $\alpha$ directly causes $\beta$. & CS/2 \\
\hline
happen_at($\tau$, $\epsilon$) & Event $\epsilon$ happens in scenario $\tau$. & HA/2 \\
\hline
real_cause($\chi$, $\epsilon$) & Observed effect $\epsilon$ is actually caused by $\chi$. & RC/2 \\
\hline
\end{tabular}
\caption{List of Predicates, Descriptions, and Notations}
\label{tab:predicates}
\end{table}

\subsubsection{Prolog Program Design}
\label{sec:prologdesign}

This step aims to synthesize common reasoning patterns and convert the schemas derived in the previous stage into \textit{Prolog} predicates that support logical verification. The procedure is outlined in Algorithm~\ref{alg:unified}. We begin by leveraging the logical implication format defined for each fallacy type in Section~\ref{sec:logicform} (Line 5). Subsequently, multiple rounds of collaborative analysis were conducted among the co-authors to extract schematic reasoning patterns associated with each fallacy type. For example, the sentence \textit{"Why do meteors always land in craters?"} exemplifies a fallacy that inverts the causal or temporal relationship between observation and explanation—thus allowing a relatively straightforward formalization. In contrast, semantically nuanced fallacies such as \textit{CT} involve distortions of quoted material or partial misinterpretations of intent, as seen in \textit{"If I continue eating an apple a day, will I never get my PhD?"}. These qualitative analyses allow us to formally capture the core structure of each fallacy (Line 6).

We then design corresponding \textit{Prolog} predicates, denoted as \texttt{pd}, incorporating both rules and example \textit{facts} to serve as few-shot prompts for the LLM (Line 7). Of the 14 fallacy types in \bench{}, 11 were selected for enhancement via this method; the remaining three exhibited sufficiently strong baseline performance (Section~\ref{sec:rq1}).

Table~\ref{tab:predicates} summarizes the full set of \texttt{pd} predicates and their semantic roles. Each predicate is carefully constructed to encode the specific reasoning flaw of its corresponding fallacy. Formally, we define $\mathcal{H}_{\tilde{F}_{\text{valid}}}$ as the set of all \textit{Prolog} fact instances satisfying the rule $\tilde{R}_\phi$ for a fallacy type $\phi$, with the mapping:
\[\small
\mathcal{H}_{\tilde{F}_{\text{valid}}} \mapstochar\longrightarrow \texttt{pd}(\texttt{argument}_1, \texttt{argument}_2, \ldots, \texttt{argument}_n)
\]
ensuring that only logically valid constructions under $\tilde{R}_\phi$ are used for natural language generation. Below, we illustrate schemas for 3 example fallacy types; full definitions for all types are provided in the appendix.

\newtheorem{mydefinition}{Definition}

\begin{mydefinition}\normalfont\bfseries
Accident Fallacy [AF]. \normalfont\mdseries This definition identifies all valid instantiations of $(O, R, I, K)$ that match the rule schema $R$-\texttt{AF}, which formalizes the accident fallacy, misinterpreting a general rule by extending it beyond its reasonable bounds. Let \texttt{O=shampoo_bottle}, \texttt{R=lather_rinse_repeat}, \texttt{I=wash_once_or_twice}, and \texttt{K=infinite_washing}. Here, $HR(O, R)$ holds since the rule appears on the shampoo bottle. A reasonable interpretation is captured by $RRI(R, I)$: the instruction implies washing once or twice. However, $RUI(R, K)$ also holds: an unreasonable interpretation would suggest one must wash infinitely. Since $I \neq K$, the conclusion formed by treating $K$ as a valid reading commits an accident fallacy. The error arises from rigidly applying a general rule without regard to practical limits or intended scope, leading to an absurd or unintended consequence.

\begin{equation}
\small
\label{eq:raf}
\frac{
\begin{aligned}
\tilde{R}_\Phi = pd(O, R, I, K) \; :- \;
& HR(O, R), \quad RRI(R, I), \\
& RUI(R, K), \quad I \neq K
\end{aligned}
}{
\mathcal{H}_{\tilde{F}_{\text{valid}}} \mapstochar\longrightarrow pd(O, R, I, K)
}
\tag{R-AF}
\end{equation}

\end{mydefinition}

\begin{mydefinition}\normalfont\bfseries
\label{def:id}
Improper Distribution or Addition [ID]. \normalfont\mdseries This definition identifies all valid tuples $(A, \Delta, E, \Upsilon)$ that instantiate the rule. Let $X$\texttt{=brush\_teeth}, A\texttt{=2\_mins}, $\Delta$\texttt{=14\_mins}, E\texttt{=teeth\_health\_for\_that\_day}, $\Upsilon$\texttt{=teeth\_health\_for\_one\_week}, and $R$\texttt{=repeat\_7\_times\_in\_one\_go}. The rule schema $R$-\texttt{ID} captures a reasoning failure where action duration can be validly accumulated, but the corresponding effect cannot. Specifically, $HE(X, A, E)$ holds since brushing for 2 minutes improves dental health for a day, and $HE(X, \Delta, \Upsilon)$ holds since brushing once for 14 minutes yields a week-long effect. Temporal accumulation is valid: $VC(A, R, \Delta)$, as seven repetitions of $A$ compose $\Delta$. However, effect-level accumulation fails: $\neg VC(E, R, \Upsilon)$. Therefore, taking the positive version $VC(E, R, \Upsilon)$ as a premise and instantiating it with such a tuple $(A, \Delta, E, \Upsilon)$ leads to a fallacy, as it incorrectly assumes that repeating short-term effects compounds into the long-term effect.

\begin{equation}
\small
\frac{
\begin{aligned}
\tilde{R}_\Phi = pd(A, \Delta, E, \Upsilon) \; :- \;
& HE(X, A, E), \quad HE(X, \Delta, \Upsilon), \\
& VC(A, R, \Delta), \quad \neg VC(E, R, \Upsilon)
\end{aligned}
}{
\mathcal{H}_{\tilde{F}_{\text{valid}}} \mapstochar\longrightarrow pd(A, \Delta, E, \Upsilon)
}
\tag{R-ID}
\end{equation}

\end{mydefinition}

\begin{mydefinition}\normalfont\bfseries
False Cause [FS].\normalfont\mdseries This definition identifies all valid instantiations of $(T, E)$ that match the rule schema $R$-\texttt{FS}, which formalizes the false cause fallacy, wrongly treating the mere temporal or spatial co-occurrence of two events as evidence that one directly produces the other as a substance. Let $T$\texttt{=lightbulb\_switch}, $E$\texttt{=darkness\_emission}, $\Upsilon$\texttt{=room\_event}, and $X$\texttt{=absence\_of\_light}. Here, $HA(\Upsilon, T)$ and $HA(\Upsilon, E)$ hold: both the action of switching the lightbulb and the resulting darkness occur as part of the same observable room event. However, $RC(X, E)$ identifies the real cause of darkness as the absence of light, not the switching action itself. Since $X \neq T$ and $T @< E$ (the switch action precedes the observed effect), the fallacy arises when one concludes that turning off a lightbulb emits darkness as a physical substance. This misrepresents a lack (the absence of illumination) as a generative act, conflating temporal correlation with causal production.

\begin{equation}
\small
\frac{
\tilde{R}_\Phi = pd(T, E) \; :- \;
\begin{aligned}
& HA(\Upsilon, T), \quad HA(\Upsilon, E), \\
& RC(X, E), \quad X \neq T, \quad T @< E
\end{aligned}
}{
\mathcal{H}_{\tilde{F}_{\text{valid}}} \mapstochar\longrightarrow pd(T, E)
} 
\tag{R-FS}
\end{equation}

\end{mydefinition}

\subsubsection{Prolog Knowledge Generation}

\label{sec:knowledgeneration}
To address the challenge of automatically generating statements containing nuanced logical fallacies, we leverage the facts and rules constructed in the previous section, combined with LLMs, to reduce human effort. Specifically, we provide the LLM with the formal rule defining the fallacy type and corresponding \textit{Prolog} facts as few-shot examples (Line 8). An important observation from our experiments is that LLMs better capture inter-predicate relationships when facts related to a specific fallacy instance are grouped together rather than by predicate name. For instance, in Equation~\ref{eq:raf}, it is more effective to group \textit{HR(O, R)}, \textit{RRI(R, I)}, and \textit{RUI(R, K)} within a single example. This grouping encourages the LLM to semantically align argument values and generate a logically coherent combination of \textit{HR}, \textit{RRI}, and \textit{RUI} predicates.

Moreover, because predicates encode relationships between arguments, adding inline comments to clarify each predicate improves the LLM’s understanding. For example, the fact \texttt{HR(highway, maximum_speed_65)} can be annotated as \texttt{\% this means highway has a rule of maximum speed 65}. These comments help LLMs more accurately infer the semantics of each predicate.
The prompt used for this task is presented in the appendix.

\subsubsection{Fallacy Sentence Transformation}
\label{sec:sentencetransfomration}
This section aims to transform appropriate fact combinations into natural language sentences that reflect the implication-style format established in Section~\ref{sec:logicform}. Specifically, we utilize the knowledge base $\mathcal{K}_{\text{prolog}}$ (Lines 9–10) and construct a query to extract all valid \textit{Prolog} knowledge facts corresponding to the rules of a given fallacy type, denoted as $\mathcal{H}_{\tilde{F}_{\text{valid}}}$ (Line 11). Finally, we instruct the LLM using both the template set $T_{\phi}$ and the retrieved facts $\mathcal{H}_{\tilde{F}_{\text{valid}}}$ to convert the logical facts into natural language sentences (Line 12). The prompt used for sentence transformation is shown in the appendix. We denote the resulting dataset as \textbf{\augbench{}}.

\section{Experimentation}
\label{sec:experiment}

\tool{} employs a \textit{Prolog}-based backend comprising 1,458 lines of code, including 24 unique fact predicates, 13 rule predicates, and 11 distinct queries. In addition, we implement 12 Python scripts (1,517 lines total) for managing LLM interaction, data preprocessing, metric computation, and visualization. Fact generation is powered by Claude 3.7 with extended thinking mode~\cite{claude37}.

Our experiments are designed to answer the following research questions:

\begin{itemize}[leftmargin=*]
\item \textbf{RQ1 (Fallacy Generation Quality):} How can we construct a logical fallacy generator that reliably produces sentences reflecting specific fallacy types?

\item \textbf{RQ2 (LLM Fallacy Detection Capability):} To what extent can LLMs detect the presence of a logical fallacy in a given sentence?

\item \textbf{RQ3 (LLM Fallacy Categorization Capability):} Can LLMs correctly categorize a fallacious sentence using the appropriate fallacy labels?
\end{itemize}

\subsection{Experiment Setup}

\subsubsection{Baseline Methods}
We compare \tool{} with two baseline methods that omit predicate engineering and structured logic definitions, relying solely on LLM internal reasoning. Details follow:

\begin{itemize}[leftmargin=*]
    \item \textbf{FallacyGen-Direct:} This baseline uses Claude 3.7~\cite{claude37}, a state-of-the-art reasoning model outperforming Deepseek-R1~\cite{deepseekr1} and GPT-o3-mini~\cite{openai2025o3mini} on benchmarks such as GPQA Diamond~\cite{anthropic2023claude}. Following a simple pipeline, we (1) compile example sentences and formal definitions for each fallacy, (2) prompt the LLM with both, and (3) instruct it to generate new instances of the same fallacy type. Variants of the prompt were explored but yielded similar results. Full prompts are detailed in the appendix.

    \item \textbf{FallacyGen-Prolog:} This method prompts LLMs to generate \textit{Prolog} programs, including facts, predicates, and inference rules, based on each fallacy’s definition and illustrative examples. The LLM is also tasked with generating corresponding natural language sentences. Unlike \tool{}, this approach imposes no structural constraints or validation mechanisms, allowing the model full control over rule construction. Details are provided in the appendix.
\end{itemize}

\subsubsection{Benchmarks and Tools}
We summarize the datasets used for each research question and the tools involved in our experiments:

\begin{itemize}[leftmargin=*]
    \item \textbf{\bench:} Used across RQ1, RQ2, and RQ3. For RQ1, it serves as the basis for validating sentence quality evaluation. For RQ2 and RQ3, it is employed to assess LLM capabilities in logical fallacy detection and categorization.

    \item \textbf{\augbench:} Also used in RQ1–RQ3. In RQ1, we reference \augbench{} as the output of \tool{}, enabling direct method comparison. In RQ2 and RQ3, it complements \bench{} to evaluate LLM performance across both synthetic and real-world data.

    \item \textbf{\soundbench:} Used in RQ2 and RQ3. We compile logically sound sentences from \textbf{C4} and \textbf{FineWeb}~\cite{allenai_c4,huggingfacefw_fineweb}, filtering for sentence tokens to match \texttt{r/ShittyAskScience}, yielding a curated set of 502 examples. This benchmark is combined with \bench{} and \augbench{} to compute metrics such as false positive (FPR) rates.

    \item \textbf{Prolog Infrastructure:} We use \textbf{SWI-Prolog}~\cite{swi-prolog}, a robust, open-source implementation of the Prolog language widely used in logic programming.
\end{itemize}

\subsubsection{Evaluated Models}
We evaluate nine state-of-the-art LLMs chosen for their relevance to logical reasoning. These models fall into two categories: \textbf{reasoning models}, which produce intermediate logic chains (e.g., DeepSeek R1~\cite{deepseekr1}, GPT-o3-mini-2025-01-31~\cite{o3mini}, Claude 3.7 with extended thinking~\cite{claude37}); and \textbf{non-reasoning models}, which do not explicitly perform stepwise reasoning (e.g., Claude 3.5~\cite{claude35}, LLaMA 3.1 405B~\cite{llama3_405b}, DeepSeek V3~\cite{deepseekv3}, GPT-4o-2024-08-06~\cite{gpt4o}, Grok-2~\cite{grok2}, Claude 3.7-20250219~\cite{claude37}). They are accessed via Cloud APIs. Full prompting templates are provided in the appendix.

\noindent

\subsection{RQ1: Fallacy Generation Quality}
\label{sec:rq1}

To systematically evaluate fallacy generation quality, we compare three methods: \textbf{FallacyGen-Direct}, \textbf{FallacyGen-Prolog}, and \textbf{\tool}. For each selected fallacy type, each method generates 20 unique sentences, producing a balanced dataset for per-fallacy comparison.

\noindent
\textbf{Sentence Quality Evaluator} 
To reduce evaluation bias, we adopt a cross-model evaluation strategy. All generations are produced using \textbf{Claude 3.7 with extended thinking}, while evaluation is performed by \textbf{GPT-4o}. This separation ensures that evaluation reflects general sentence quality, independent of the generation model. Each sentence is scored on a 0--3 scale, with 3 indicating strong alignment with the intended fallacy. All evaluations are run at temperature 0, and each sentence is scored three times. Evaluation prompt details are provided in the appendix.

\begin{figure}[!ht]
    \centering
    \includegraphics[width=1\linewidth]{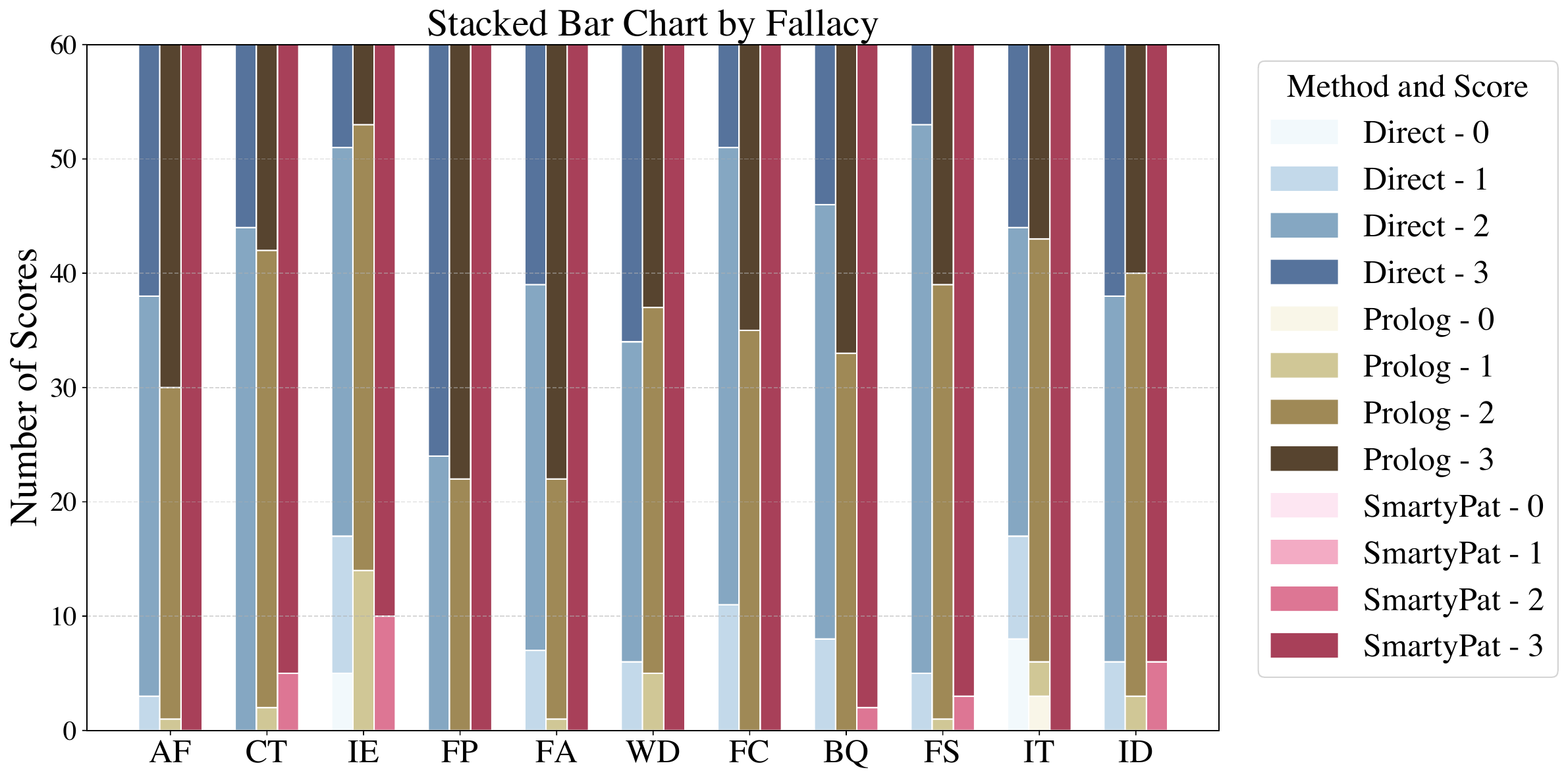}
    \caption{The score distribution of the three methods across different types of logical fallacies. \footnotesize{*Direct means \textbf{FallacyGen-Direct}, Prolog means \textbf{FallacyGen-Direct}.More score 3 is better.}}
    \label{fig:scores_statistics}
\end{figure}

\noindent
\textbf{Sentence Quality Evaluator Validity.} 
We first validated the evaluator’s effectiveness on the original \bench{}, and further confirmed its consistency with human annotations. 
Details of the prompts, annotation process, and agreement statistics are provided in the Appendix.

\noindent
\textbf{Baseline Testing and Analysis.} FallacyGen-Direct excels at generating fallacies reliant on surface-level features, such as \textit{EC}, \textit{NF}, and \textit{FD}. These fallacies often rely on shallow lexical ambiguity or binary framing, patterns LLMs can easily mimic without deep reasoning. For example, \textit{EC} and \textit{NF} exploit ambiguity in terms like \textit{pound} or phrases like \textit{burn calories}, while \textit{FD} leverages binary constructions. This highlights that LLMs perform well on linguistically superficial fallacies but struggle with semantically complex ones. In contrast, fallacies like \textit{CT} require context or sociocultural awareness, posing greater difficulty. We therefore exclude these trivial fallacies from Prolog-based generation, as FallacyGen-Direct already performs near optimally on them.

\noindent
\textbf{\tool Effectiveness.} As shown in Figure~\ref{fig:scores_statistics}, \tool{} significantly reduces low-quality outputs (scores 0--1) and increases high-quality outputs (scores 2--3) across nearly all fallacy types. Notably, for \textit{FC}, \tool{} generates 60 score-3 instances, outperforming FallacyGen-Prolog (25) and FallacyGen-Direct (9), demonstrating superior semantic and structural alignment. Additionally, semantic similarity analysis using \texttt{text-embedding-3-large}~\cite{openai2024embedding} shows intra- and inter-benchmark cosine similarity of ~0.16, confirming the novelty of generated content. Full results and embeddings are detailed in the appendix.

\noindent
\textbf{Comparison of Average Scores.} We report average sentence quality scores by fallacy type across methods; the full table is provided in the appendix. \tool{} consistently achieves the highest average score in all categories. The ``Enhance'' row quantifies the relative improvement of \tool{} over FallacyGen-Direct. On average, \tool{} outperforms this baseline by 38.12\%, with particularly large gains observed in \textit{\textbf{IE} (\textbf{+58.88\%}), \textbf{IT} (\textbf{+62.16\%}), \textbf{FC} (\textbf{+52.54\%}), and \textbf{FS} (\textbf{+45.08\%})}. These results highlight \tool{}'s advantage in enhancing both structural accuracy and semantic fidelity, demonstrating its effectiveness for generating high-quality, logic-driven fallacy instances.

\begin{figure}[!ht]
    \centering
    \includegraphics[width=0.49\linewidth]{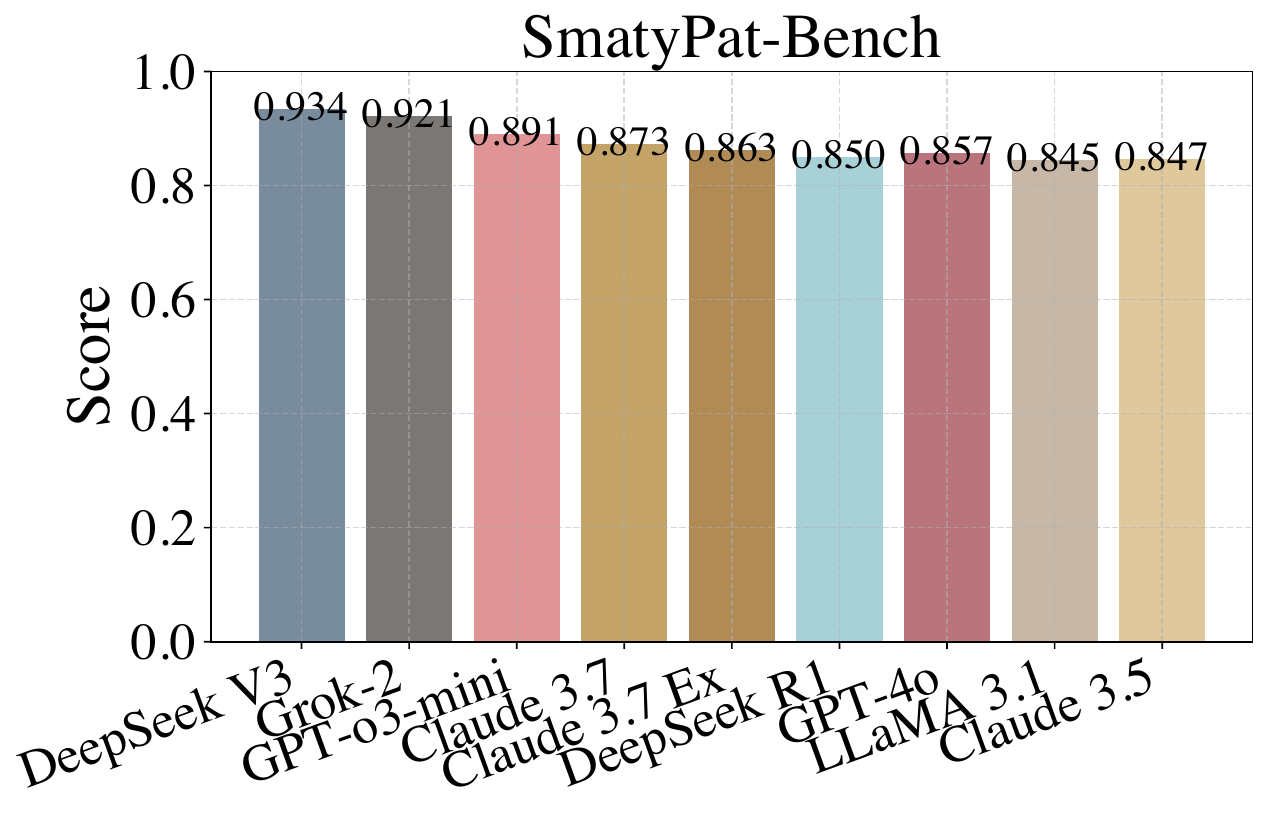}
    \includegraphics[width=0.49\linewidth]{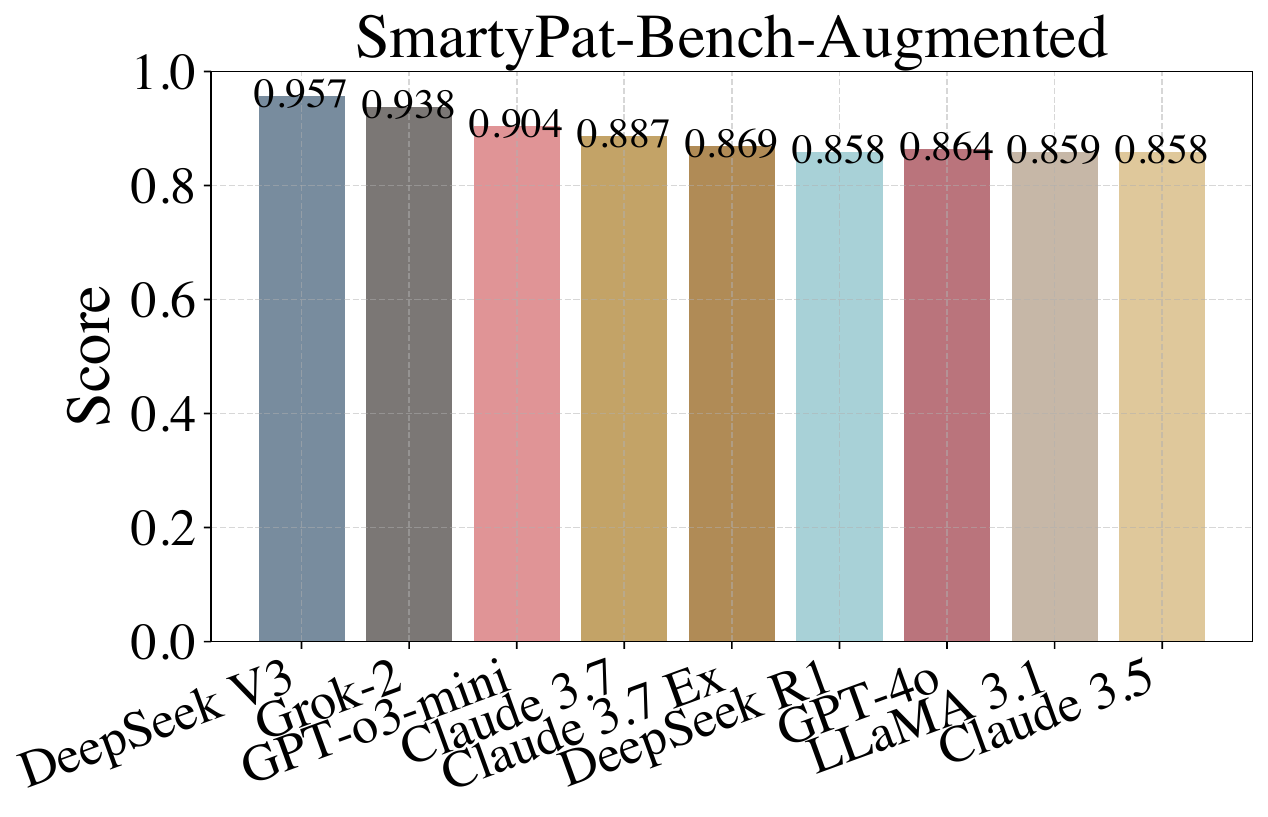}
    \caption{F1 score(higher better), sorted by F1 score in descending order.  \footnotesize{Claude 3.7 Ex means Claude 3.7 Extended Thinking.} } 
    \label{fig:fp_fn_f1}
\end{figure}

\subsection{RQ2: LLM Fallacy Existence Detection Capability}

We evaluate nine state-of-the-art LLMs on their ability to detect the presence of logical fallacies in sentences. Our analysis addresses two key dimensions: overall detection performance and fallacy-specific detection difficulty.

\noindent
\textbf{Logical Fallacy Detection Ability} We report \textit{FPR}, where logically sound sentences are misclassified as fallacious; \textit{FNR}, where fallacies are missed; and the resulting F1-score. Figure~\ref{fig:fp_fn_f1} presents the F1 scores, with the complete figure available in the appendix. \augbench{} exhibits FPR/FNR/F1 patterns comparable to \bench{}, indicating LLMs perceive both datasets as similarly fallacious. Interestingly, non-reasoning models (e.g., DeepSeek V3, Grok-2) consistently outperform reasoning models (e.g., Claude 3.7, GPT-o3-mini) in F1-score, despite expectations to the contrary. Closer inspection suggests reasoning models tend to overanalyze, flagging benign content as fallacious. For instance, Claude 3.7 mislabels a simple instructional sentence (`If you are a beginner, it is best to begin with a flat board') as an \textit{AF} due to overgeneralization. This behavior likely reflects a form of \textit{confirmation bias}~\cite{o2025confirmation}, where models assume fallacies must be present. All models show high FPR and near-zero FNR, with reasoning models generally more prone to false positives. \textbf{These findings reveal a sensitivity bias in LLMs, with reasoning models not outperforming non-reasoning models in detection accuracy.}

\begin{figure}[!ht]
    \centering
    \includegraphics[width=1\linewidth]{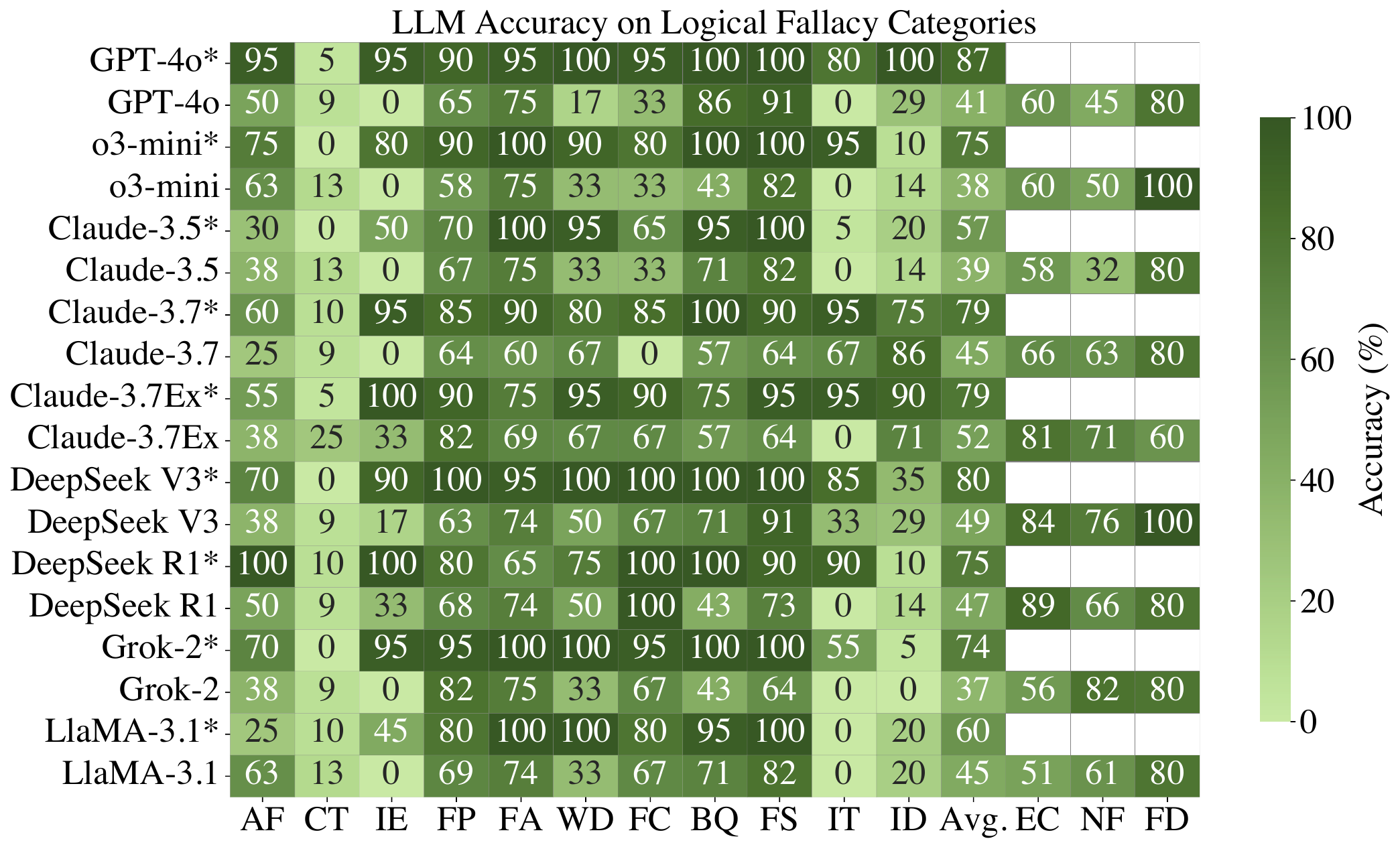}
    \caption{LLM accuracy (darker is better) in identifying fallacies from \bench{} (without *) and \augbench{} (with *).}
    \label{tab:accuracy}
\end{figure}

\noindent
\textbf{Fallacy-wise Detection Ability.} Using \augbench{}, we compute fallacy-wise detection accuracy (Figure~\ref{tab:accuracy}) by checking whether the ground-truth fallacy label appears in the model’s predictions. Fallacy types in \augbench{} are notably easier to identify. Averaged across models, LLMs perform best on \textit{FS} and \textit{FA}, consistent with intuitive causal and analogical reasoning. In contrast, detection rates are lower for context-dependent fallacies such as \textit{CT}, \textit{IT}, and \textit{ID}. \textbf{These results suggest that LLMs handle surface-level causal or analogical fallacies well but struggle with those requiring nuanced contextual understanding.}

\subsection{RQ3: LLM Fallacy Categorization Capability}
This section evaluates whether LLMs can accurately assign fallacy labels to given sentences. We test nine LLMs on both \bench{} and \augbench{}. Unlike the detection task (Figure~\ref{tab:accuracy}), this task assesses the overall similarity between predicted and ground-truth label sets, factoring in prediction rank and the presence of incorrect labels.

\noindent
\textbf{Ranked Fallacy Scorer.} We adopt a rank-weighted scoring function. Let \( G = [g_1, g_2, \dots, g_m] \) be the ground-truth labels and \( P = [p_1, p_2, \dots, p_n] \) the predicted labels. The score is calculated by:

{\small
\[
S_{\text{original}}(G, P) = \sum_{i=1}^{n} 
\left\{
\begin{array}{ll}
\frac{1}{i}, & \text{if } p_i \in G \\[3pt]
-\frac{1}{i}, & \text{if } p_i \notin G
\end{array}
\right.
\]
}

For instance, a correct top-ranked prediction yields \(+1\), while an incorrect one yields \(-1\). The worst-case score occurs when none of the predicted labels appear in \( G \), particularly when predicting nearly all possible labels (\(T-1\), where \(T\) is the total number of fallacy types), leading to a penalty of \(-\sum_{i=1}^{T-1} \frac{1}{i}\).

\begin{figure}[!ht]
    \centering
    \begin{subfigure}[t]{1\linewidth}
        \centering
        \includegraphics[width=\linewidth]{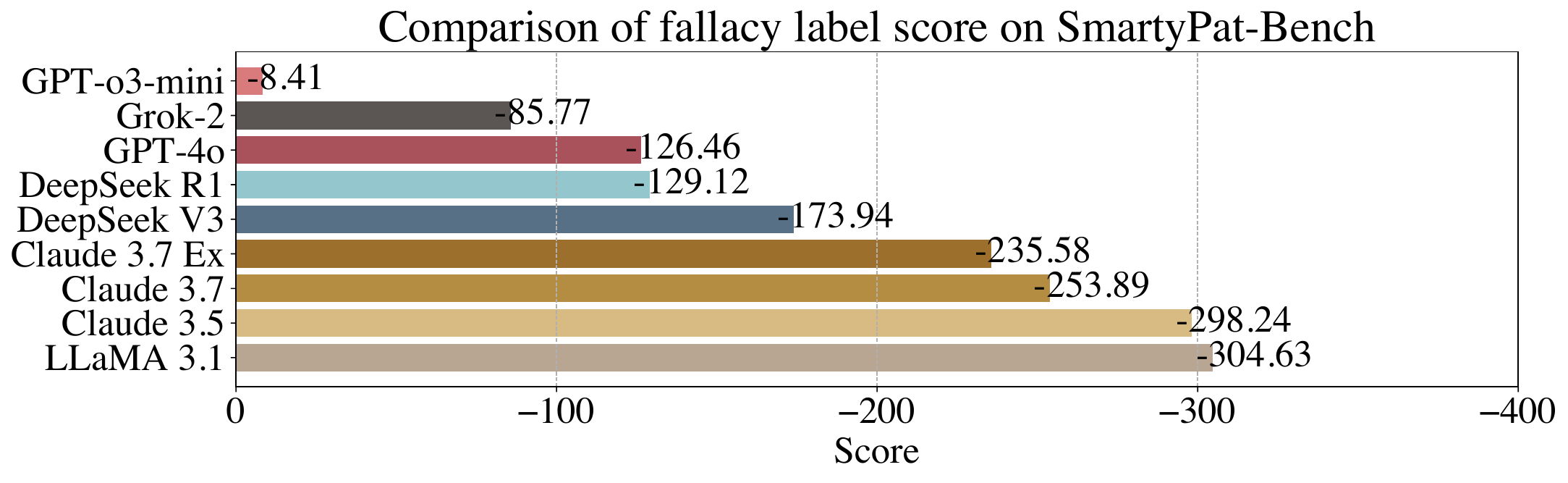}
    \end{subfigure}
    \hfill
    \begin{subfigure}[t]{1\linewidth}
        \centering
        \includegraphics[width=\linewidth]{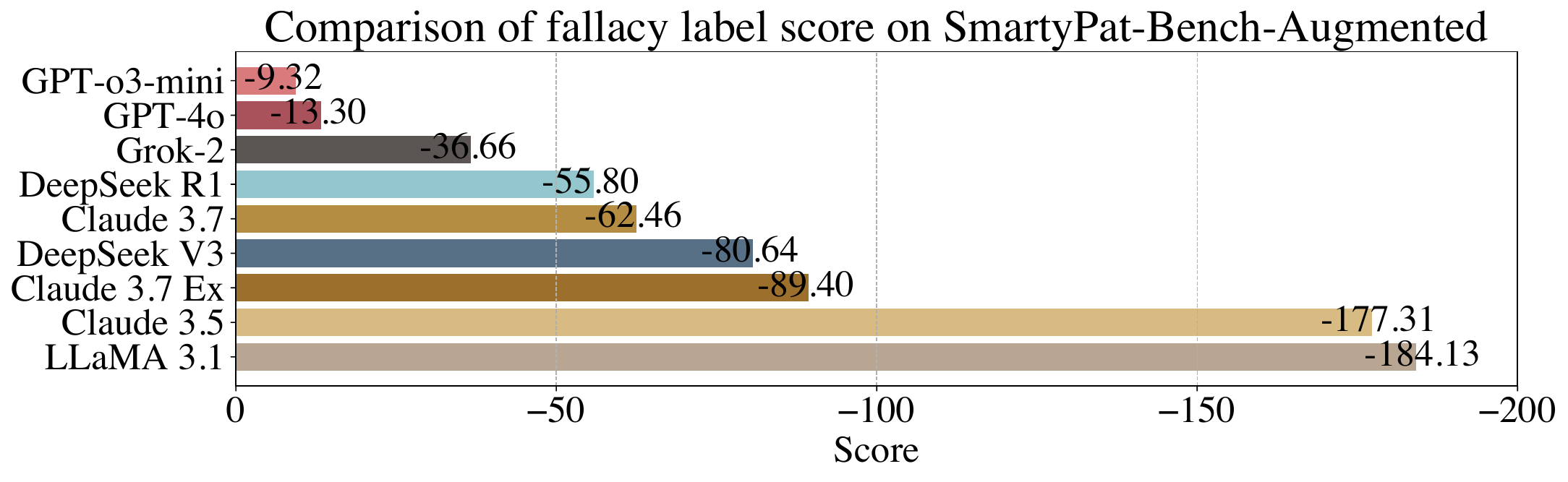}
    \end{subfigure}
    \caption{Fallacy label scores for selected LLMs, sorted in descending order (Close to the Left is better).}
    \label{fig:combined_fallacy_graphs}
\end{figure}

\noindent
\textbf{Fallacy Categorization Capabilities.} Figure~\ref{fig:combined_fallacy_graphs} shows model scores on \bench{} (top) and \augbench{} (bottom). Scores on \augbench{} are consistently higher, reflecting more explicit fallacies structures. Claude 3.7 and its extended version improve by over 70\%, likely due to alignment with familiar generation patterns~\cite{panickssery2024llm}. Reasoning models generally outperform non-reasoning ones; notably, GPT-o3-mini surpasses GPT-4o, and DeepSeek R1 exceeds DeepSeek V3. \textbf{Overall, reasoning-oriented models are better suited for fallacy classification.}

We also evaluate Grok-2’s behavior (see Appendix for details, limitations, and future work). Its unexpectedly strong performance arises from a conservative labeling strategy that minimizes penalties. Moreover, the sentences generated by \tool{} exhibit clearer and more explicit fallacy structures, illustrating how the integration of LLMs with formal methods can enhance output stability and enable verifiable generation, which is an emerging direction toward the trustworthy deployment of LLMs in critical applications.

\section{Conclusion}

We propose \bench{}, the first real-world native English dataset of fallacious questions, and \tool{}, a novel Prolog-based method for generating high-quality fallacy sentences, resulting in \augbench{}. \tool{} outperforms two baselines in generating accurate, diverse fallacies. Evaluating nine state-of-the-art LLMs, we find that stronger models tend to overanalyze, causing high false positives. Non-reasoning models excel at detection, while reasoning models perform better at categorization. The GPT series, especially GPT-o3-mini, offers the best overall balance across both tasks.

\bibliography{aaai2026}

\section{APPENDIX}
\subsection{Introduction Appendix}
\begin{figure*}[!ht]
    \centering
    \includegraphics[width=1\linewidth]{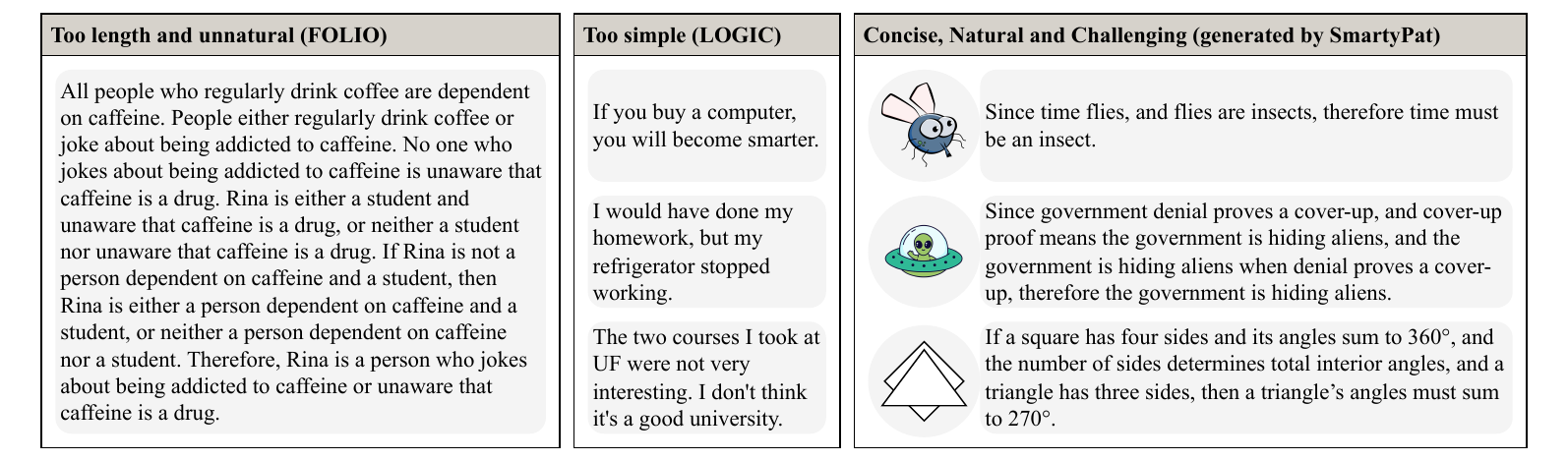}
    \caption{Examples of testcases in different benchmarks.\footnotesize{*Each grey-colored block is a single testcase.}}
    \label{fig:bench_cmp}
\end{figure*}

Examples of testcases in different benchmarks can be found in Table \ref{fig:bench_cmp}.

\subsection{Related Work Appendix}

\head{Logic Reasoning Benchmarking}
Traditional evaluation methods assess various logical inference forms, including inductive reasoning, which involves drawing conclusions from a set of observed propositions and generalizing patterns from specific cases~\cite{sinha2019clutrr}; deductive reasoning, which derives logically certain conclusions from all available observations, even in the presence of unobserved special cases~\cite{tian2021diagnosing}; and abductive reasoning, which aims to identify the most plausible explanation for a given set of observations~\cite{del2022true}. These reasoning tasks typically require constructing complex first-order logic representations that involve the use of quantifiers (\(\forall\) for universal quantification, \(\exists\) for existential quantification, \(\exists!\) for unique existence, and \(\nexists\) for negated existence), as well as logical connectives such as conjunction (\(\wedge\)), disjunction (\(\vee\)), negation (\(\neg\)), implication (\(\rightarrow\)), biconditional (\(\leftrightarrow\)), exclusive or (\(\oplus\)), nand (\(\uparrow\)), and nor (\(\downarrow\)). To rigorously assess an LLM’s inferential capabilities, benchmarks such as FOLIO~\cite{han2022folio} and LOGICBench~\cite{parmar2024logicbench} provide structured evaluation settings that test its ability to navigate these logical constructs effectively.

\subsubsection{Logical Fallacy Categorization Details}

Table~\ref{tab:fallacy_labels} shows the details about the fallacy categorization proposed by us.
The examples are the raw Reddit posts we collected for \bench{}. To ensure the reliability and consistency of our definitions while simplifying the labeling process, we referred to two authoritative logic textbooks \cite{gula2002nonsense, bennett2012logically}. 
\begin{table*}[!htb]
\centering
\renewcommand{\arraystretch}{1.3}
\caption{Detailed definitions and examples of logical fallacies}
\setlength{\tabcolsep}{6pt} 
\arrayrulewidth=1pt
\rowcolors{2}{gray!15}{white} 
\resizebox{\textwidth}{!}{%
\begin{tabular}{m{3cm} m{7.5cm} m{6.5cm} m{1.3cm}}
\hline
\rowcolor{gray!30} 
\textbf{Categorisation} & \textbf{Definition} & \textbf{Example} & \textbf{Notation} \\ \hline
\textbf{False Dilemma} & The presentation of an issue as having only two possible outcomes, either right or wrong, without recognising that additional alternatives may exist. & Where can I buy the toothpaste that only 1 out of 5 dentists recommends? & FD \\ \hline
\textbf{Equivocation} & The misleading use of a word or phrase that has multiple meanings, creating ambiguity and leading to confusion in interpretation or reasoning. & If smoking is so bad for you, how come it cures salmon? & EC \\ \hline
\textbf{False Premise} & The establishment of an argument based on an unfounded, non-existent, or unreasonable assumption, leading to flawed reasoning or invalid conclusions. & How did Thomas Edison come up with the idea for the lightbulb if the lightbulb didn't exist to appear above his head? & FP \\ \hline
\textbf{False Analogy} & The assumption that if A and B share certain characteristics, then B must also possess other attributes of A, despite lacking a valid basis for this inference. & If coconuts have hair and produce milk, why aren't they mammals? & FA \\ \hline
\textbf{Wrong Direction} & The incorrect attribution of causality by reversing the cause-and-effect relationship, assuming the effect is the cause and the cause is the effect. & Why do meteors always land in craters? & WD \\ \hline
\textbf{Fallacy of Composition} & The mistaken assumption that what is true for a part of something must also be true for the whole, disregarding the possible differences between individual components and the entire entity. & If seat belts are so safe, why don't they just make cars out of seat belts? & FC \\ \hline
\textbf{Begging the Question} & The use of a statement as both the premise and the conclusion, assuming the truth of what is to be proven instead of providing independent support. & If people hate spoilers, then why did Snape kill Dumbledore? & BQ \\ \hline
\textbf{False Cause} & The incorrect assumption that a causal relationship exists between two events solely because one follows the other, failing to account for coincidence or other influencing factors. & When I drink alcohol, I feel great. The next day when I drink water, I feel terrible. Why is water so bad for you? & FS \\ \hline
\textbf{Inverse Error} & The mistaken reasoning that if A implies B, then not A must imply not B, overlooking the possibility that B may still occur due to other factors. & If I pedal backwards on my exercise bike, will I gain weight? & IE \\ \hline
\textbf{Improper Transposition} & The incorrect inference that if A implies B, then B must also imply A, failing to recognise that implication is not necessarily reversible. & If a picture is worth one thousand words, how many is a picture of one word? & IT \\ \hline
\textbf{Improper Distribution or Addition} & The erroneous reasoning that individual effects can be directly summed or distributed across a group without considering their actual impact or interaction. & If I brush my teeth for 28 minutes once a week instead of two minutes twice a week, will the effect still be the same? & ID \\ \hline
\textbf{Contextomy} & The act of selectively quoting or altering a statement, advertisement, or published material in a way that distorts its original meaning, often misrepresenting the intent of the original source. & My dad told me I have to hold my breath when I drive through a tunnel. But now I'm in a tunnel and traffic is stopped. What do I do? (Time-sensitive inquiry) & CT \\ \hline
\textbf{Nominal Fallacy} & The mistaken interpretation of a metaphorical or figurative expression as a literal statement, leading to a misunderstanding of its intended meaning. & My dad said that the world doesn't revolve around me. How is this possible if I am his sun? & NF \\ \hline
\textbf{Accident Fallacy} & The misapplication of a general rule to a specific case where exceptions should be considered, treating the rule as absolute without regard for context or relevant circumstances. & If E=MC², why isn't Elephant spelled MC²LMC²phant? & AF \\ \hline
\end{tabular}%
}
\label{tab:fallacy_labels}
\end{table*}

\subsubsection{\bench{} Details}

\begin{figure}[!hbt]
    \centering
    \includegraphics[width=1.0\linewidth]{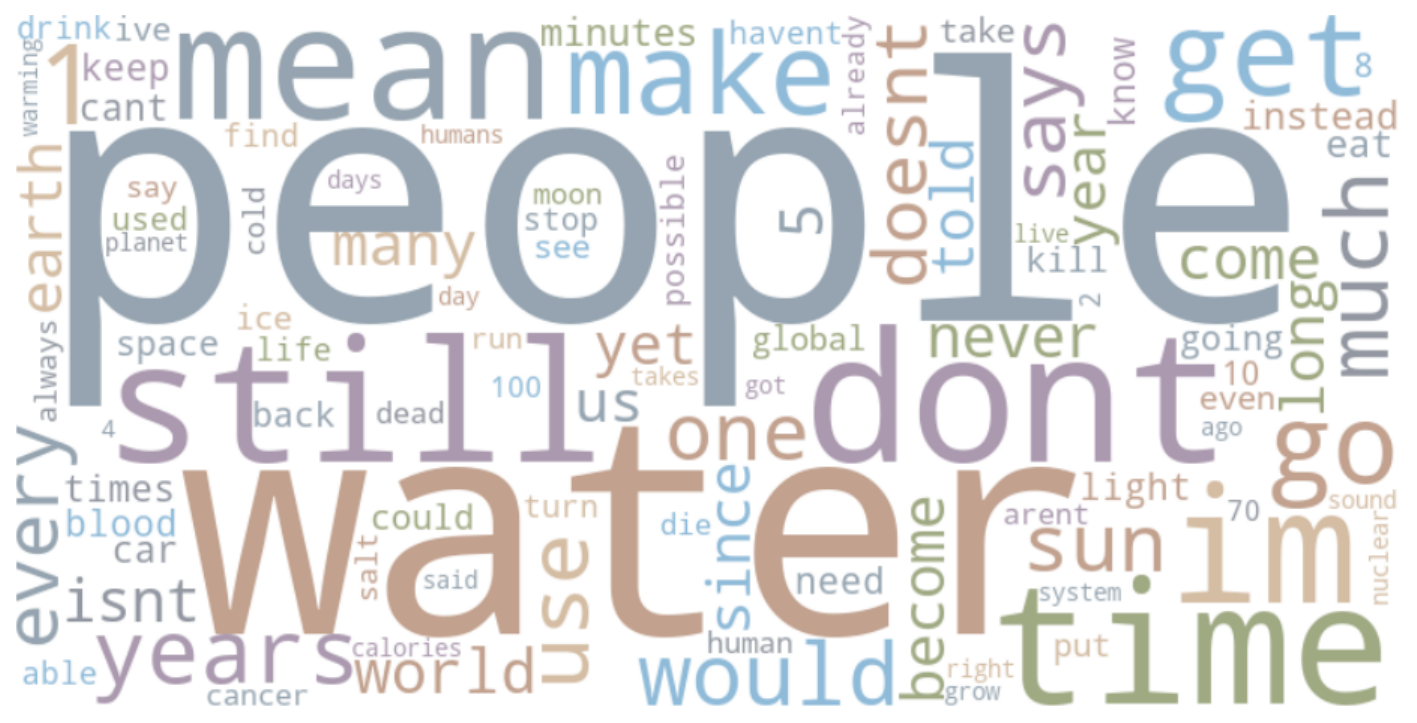}
    \caption{Word cloud visualization of the dataset. Larger words indicate higher frequency.}
    \label{fig:wordcloud}
\end{figure}

Figure~\ref{fig:wordcloud} presents the word cloud distribution of our dataset. The size of each word represents its frequency, with larger words appearing more often in the dataset. Common words include \textit{people, water, time, get, make, mean}, suggesting that discussions frequently revolve around human-related concepts, scientific phenomena, and logical reasoning.

\subsubsection{Logic Programming}

Logic programming is a programming paradigm based on formal logic.
Some famous logic programming languages are Prolog, Answer Set Programming (ASP) and Datalog.
In this paper, we focus on Prolog.
Prolog was originally developed to support artificial intelligence~\cite{bobrow1985if, rowe1988artificial}, particularly in natural language processing applications~\cite{nugues2006introduction}.
Prolog programs are made up of \textit{facts} and \textit{rules}.
Users can use \textit{queries} to ask Prolog to evaluate certain statements based on the facts and rules.
Here we provide definitions of these key concepts:

\head{Facts.}
Facts represent the basic assertions about the world.
They state what is \textit{unconditionally true}.
A fact is made up of a $predicate$ and several $entities$, denoted as:
\begin{equation}
\begin{aligned}
predicate(entity_1, entity_2, ...)
\end{aligned}
\label{eq:fact}
\end{equation}
An example is $father(john, mary)$, which means John is the father of Mary.

\head{Rules.} 
Rules are the logical relationships between facts and/or other rules.
They are the basis for inferring new knowledge.
A rule is made up of a $head$ (a new predicate) and a $body$ (a sequence of facts or rules separated by commas), denoted as:
\begin{equation}
\begin{aligned}
predicate'(...):- \;predicate_1(...), predicate_2(...), ...
\end{aligned}
\label{eq:rule}
\end{equation}
An example is $parent(X,Y):- \;father(X,Y)$, which means X is a parent of Y if X is the father of Y.
Another example is $grandparent(X,Z):- \;parent(X,Y),parent(Y,Z)$, which means that X is a grandparent of Z if X is a parent of Y and Y is a parent of Z.

\head{Queries.}
Queries have the same structure as the body of rules, denoted as:
\begin{equation}
\begin{aligned}
?- \;predicate_1(...), predicate_2(...), ...
\end{aligned}
\label{eq:query}
\end{equation}
The logic reasoning engine can provide three types of responds to the queries.
If the engine can prove the statement of the query, then it will return $True$.
If it cannot prove, it will return $False$.
If the query contains variables, then the engine can return variable bindings.
For example, given the query $?-\;parent(X, mary)$, the engine will return $X=john$.

\head{Reasoning Rules.}
Generating new facts through logic programming requires facts (\ref{eq:fact}), rules (\ref{eq:rule}), queries (\ref{eq:query}), and answers to the queries.
In some papers~\cite{drowzee, ccs-logic}, the authors simplify this process as \textit{reasoning rules} denoted as follows:
\begin{equation}
\begin{aligned}
\frac{
        \begin{array}{c}
          predicate_1(...), predicate_2(...), ...\\
        \end{array}
    }{
    \begin{array}{c}
     conclusion(...)  
      \end{array} 
      } \quad [rule\;name].
\end{aligned}
\label{eq:reasoning rules}
\end{equation}

\subsection{A Preliminary Study with SMARTYPAT-BENCH Appendix}
\subsubsection{\tool Workflow}
\begin{figure*}[!hbt]
    \centering
    \includegraphics[width=\textwidth]{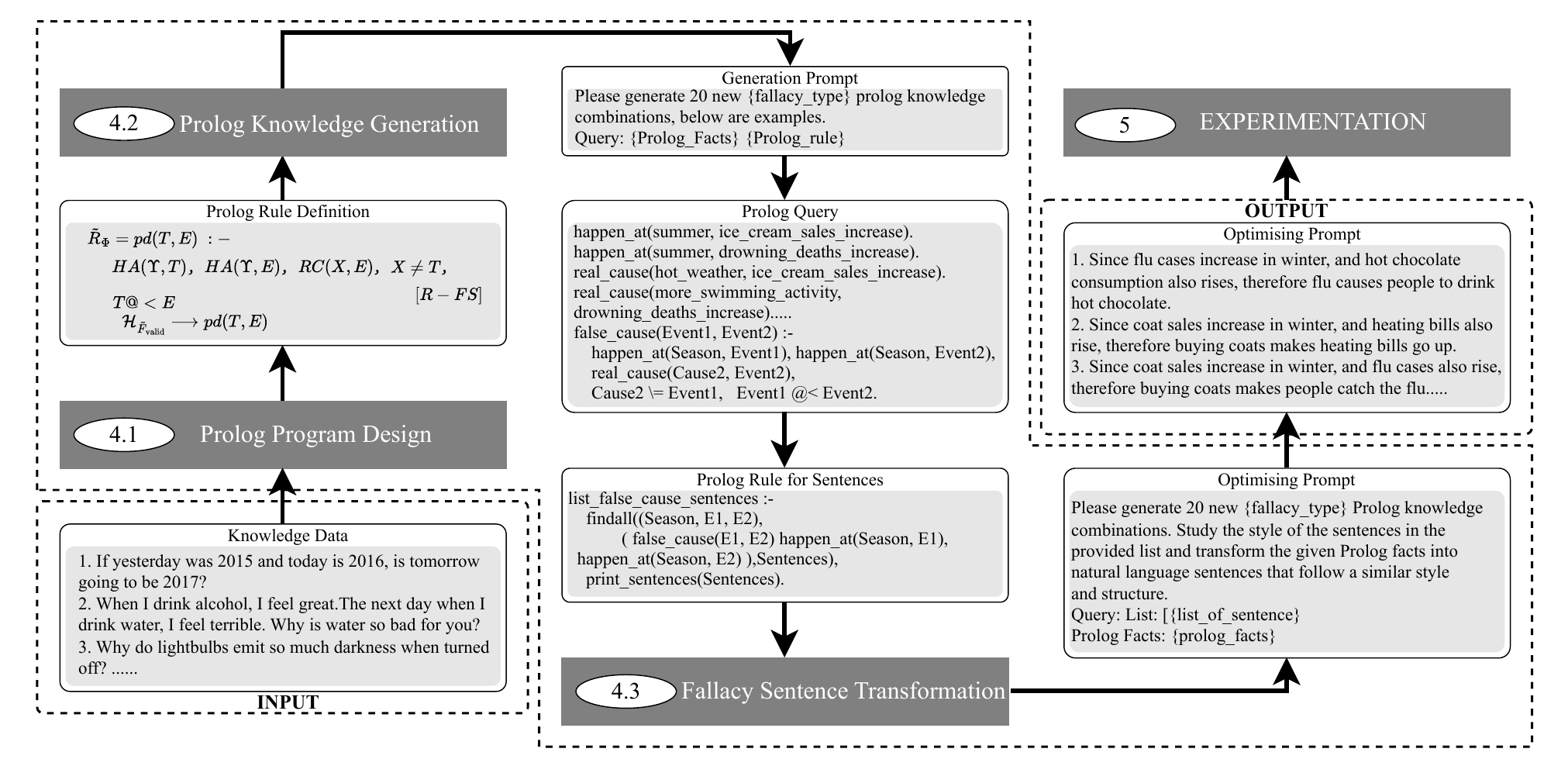}
    \caption{The workflow of \tool{}. (Take \textit{False Cause} as an example.)}
    \label{fig:prologworkflow}
\end{figure*}

The Figure \ref{fig:prologworkflow} shows the workflow of our method.

\subsubsection{Construction of \bench{}}
Inspired by the development of the \ruozhiba{} benchmark, we searched English-language forums for suitable sources of logical fallacies. 
We identified a subreddit~\cite{reddit_shittyaskscience} that could serve as an English counterpart to the Chinese forum \ruozhiba{}. 
However, posts from this subreddit could not directly serve as a benchmark and we took three steps to build the benchmark.

\subsubsection{Data Cleaning}
As of 2024, Reddit has restricted search engines from crawling its data~\cite{Emma2024}, but the Arctic Shift dataset~\cite{heitmann_arctic_shift} provides access to raw Reddit content. Using this resource, we manually extracted all posts from the target subreddit up to December 2024, creating an initial dataset of 251,052 entries. To efficiently construct a high-quality \bench{} dataset, we developed a structured preprocessing pipeline: we first performed \textit{keyword filtering} to exclude inappropriate or unethical content, then applied an \textit{upvote-based selection} strategy to identify high-quality posts by selecting the top 2,500 entries with the most engagement. Finally, five expert annotators manually reviewed these entries to confirm logical fallacies and appropriateness for public use. This procedure resulted in a linguistically diverse, reliable \bench{} dataset containing 502 logically flawed posts.

\subsubsection{Fallacy Labeling}
To ensure annotation quality and consistency, five authors independently labeled the dataset by selecting all applicable fallacy types from the 14 predefined categories. The annotation process achieved a high level of agreement, with a Fleiss’ Kappa score exceeding 0.8. Subsequently, discrepancies were discussed and resolved through joint verification to reach consensus. 
Each sentence could receive multiple fallacy labels; for instance, \textit{"Our doctor said that my wife and I are going to have a sun. How can I harness its extensive energy when my wife gives birth?"} is annotated with both \textit{Equivocation} and \textit{False Analogy}.

\subsubsection{Sentence Transformation}\label{sec:logicform}
This step standardizes the syntactic and logical structure of sentences in \bench{} for further analysis. The original dataset consists mainly of questions (e.g., \textit{"Why do meteors always land in craters?"}), often embedding implicit premises and lacking explicit inferential structure. We represent sentences using first-order logic, constrained to conjunctions ($\land$) and implications ($\rightarrow$), sufficient for modeling typical fallacious reasoning. Formally, sentences are normalized to $(p_1 \land p_2 \land \dots \land p_n) \rightarrow q$, where each $p_i$ is an atomic premise and $q$ is the conclusion. To align with natural language, paraphrastic forms such as \textit{"Since $p_1$ and $p_2$, therefore $q$"} or \textit{"If $p_1$ and $p_2$, then $q$"} are employed. For example, the question \textit{"Why do meteors always land in craters?"} is transformed into \textit{"Since we always find meteors in craters, therefore craters cause meteors."}, explicitly highlighting the flawed inference. This normalisation was manually performed by five authors, ensuring careful semantic interpretation and logical accuracy.

\subsection{Methodology Appendix}
\subsubsection{A Complete List of Fallacy Definitions}

\begin{mydefinition}\normalfont\bfseries
\label{def:id}
Improper Distribution or Addition [ID]. \normalfont\mdseries This definition identifies all valid tuples $(A, \Delta, E, \Upsilon)$ that instantiate the rule. Let $X = \texttt{brush\_teeth}$, $A = \texttt{2\_mins}$, $\Delta = \texttt{14\_mins}$, $E = \texttt{teeth\_health\_for\_that\_day}$, $\Upsilon = \texttt{teeth\_health\_for\_one\_week}$, and $R = \texttt{repeat\_7\_times\_in\_one\_go}$. The rule schema $R$-\texttt{ID} captures a reasoning failure where action duration can be validly accumulated, but the corresponding effect cannot. Specifically, $HE(X, A, E)$ holds since brushing for 2 minutes improves dental health for a day, and $HE(X, \Delta, \Upsilon)$ holds since brushing once for 14 minutes yields a week-long effect. Temporal accumulation is valid: $VC(A, R, \Delta)$, as seven repetitions of $A$ compose $\Delta$. However, effect-level accumulation fails: $\neg VC(E, R, \Upsilon)$. Therefore, taking the positive version $VC(E, R, \Upsilon)$ as a premise and instantiating it with such a tuple $(A, \Delta, E, \Upsilon)$ leads to a fallacy, as it incorrectly assumes that repeating short-term effects compounds into the long-term effect.

\begin{equation}
\small
\frac{
\begin{aligned}
\tilde{R}_\Phi = pd(A, \Delta, E, \Upsilon) \; :- \;
& HE(X, A, E), \quad HE(X, \Delta, \Upsilon), \\
& VC(A, R, \Delta), \quad \neg VC(E, R, \Upsilon)
\end{aligned}
}{
\mathcal{H}_{\tilde{F}_{\text{valid}}} \mapstochar\longrightarrow pd(A, \Delta, E, \Upsilon)
}
\tag{R-ID}
\end{equation}

\end{mydefinition}

\begin{mydefinition}\normalfont\bfseries
\textbf{False Analogy [FA].} \normalfont\mdseries
This definition identifies all valid instantiations of $(E, \Pi, X, \Phi)$ that match the rule structure. Let $E = \texttt{kid}$, $\Pi = \texttt{kidney}$, $X = \texttt{kid\_word}$, and $\Phi = \texttt{grow\_into\_adult}$. The rule schema $R$-\texttt{FA} captures a false analogy pattern, where two expressions share a lexical substructure but diverge semantically. Concretely, $HP(E, X)$ and $HP(\Pi, X)$ hold since both \texttt{kid} and \texttt{kidney} has the property substring \texttt{kid\_word}. Moreover, $HP(E, \Phi)$ holds, as the concept \texttt{kid} plausibly relates to \texttt{grow\_into\_adult}. However, $E \neq \Pi$ and $\neg HP(\Pi, \Phi)$—the \texttt{kid} in \texttt{kidney} does not grow into an \texttt{adult-ney}—so inferring $HP(\Pi, \Phi)$ based solely on shared morphology constitutes a category mistake. The fallacy arises by overextending a single property sharing ($X$) to imply a general case property sharing ($\Phi$), which does not hold.

\begin{equation}
\small
\frac{
\begin{aligned}
\tilde{R}_\Phi = pd(E, \Pi, X, \Phi) \; :- \;
& HP(E, X), \quad HP(\Pi, X), \\
& HP(E, \Phi), \quad E \neq \Pi, \quad \neg HP(\Pi, \Phi)
\end{aligned}
}{
\mathcal{H}_{\tilde{F}_{\text{valid}}} \mapstochar\longrightarrow pd(E, \Pi, X, \Phi)
}
\tag{R-FA}
\end{equation}

\end{mydefinition}

\begin{mydefinition}\normalfont\bfseries
False Premise [FP]. \normalfont\mdseries This definition identifies all valid instantiations of $(X, \Phi, \Pi, O, \Gamma)$ matching the structure of rule schema $R$-\texttt{FP}, which captures reasoning based on a false premise propagated through a plausible observation. Let $X = \texttt{people\_has\_two\_lungs}$, $\Phi = \texttt{two\_lungs\_breathe\_out\_carbon\_dioxide}$, $\Pi = \texttt{lung\_number\_influence\_carbon\_number}$, $O = \texttt{people\_can\_have\_one\_lung}$, and $\Gamma = \texttt{one\_lung\_breathe\_out\_carbon\_monoxide}$. Here, $EF(X, \Phi)$ holds: the established fact that people have two lungs and that two lungs breathe out carbon dioxide. A false premise is introduced via $FP(\Phi, \Pi)$, incorrectly claiming that the number of lungs determines the type of carbon compound exhaled. Then, $PO(O, \Pi)$ holds, since it is plausible to observe that some people have only one lung, and this could appear to support $\Pi$. Finally, $FPLC(\Pi, O, \Gamma)$ concludes that one lung leads to exhalation of carbon monoxide. The fallacy arises from accepting $\Pi$—a false causal relationship—as a valid bridge between an observation and a conclusion, thus generating $\Gamma$ from a structurally valid but semantically invalid reasoning chain.

\begin{equation}
\small
\frac{
\begin{aligned}
\tilde{R}_\Phi = pd(X, \Phi, \Pi, O, \Gamma) \; :- \;
& EF(X, \Phi), \quad FP(\Phi, \Pi), \\
& PO(O, \Pi), \quad FPLC(\Pi, O, \Gamma)
\end{aligned}
}{
\mathcal{H}_{\tilde{F}_{\text{valid}}} \mapstochar\longrightarrow pd(X, \Phi, \Pi, O, \Gamma)
}
\tag{R-FP}
\end{equation}

\end{mydefinition}

\begin{mydefinition}\normalfont\bfseries
Accident Fallacy [AF]. \normalfont\mdseries This definition identifies all valid instantiations of $(O, R, I, K)$ that match the rule schema $R$-\texttt{AF}, which formalizes the accident fallacy—misinterpreting a general rule by extending it beyond its reasonable bounds. Let $O = \texttt{shampoo\_bottle}$, $R = \texttt{lather\_rinse\_repeat}$, $I = \texttt{wash\_once\_or\_twice}$, and $K = \texttt{infinite\_washing}$. Here, $HR(O, R)$ holds since the rule appears on the shampoo bottle. A reasonable interpretation is captured by $RRI(R, I)$: the instruction implies washing once or twice. However, $RUI(R, K)$ also holds: an unreasonable interpretation would suggest one must wash infinitely. Since $I \neq K$, the conclusion formed by treating $K$ as a valid reading commits an accident fallacy. The error arises from rigidly applying a general rule without regard to practical limits or intended scope, leading to an absurd or unintended consequence.

\begin{equation}
\small
\label{eq:raf}
\frac{
\begin{aligned}
\tilde{R}_\Phi = pd(O, R, I, K) \; :- \;
& HR(O, R), \quad RRI(R, I), \\
& RUI(R, K), \quad I \neq K
\end{aligned}
}{
\mathcal{H}_{\tilde{F}_{\text{valid}}} \mapstochar\longrightarrow pd(O, R, I, K)
}
\tag{R-AF}
\end{equation}

\end{mydefinition}

\begin{mydefinition}\normalfont\bfseries
Fallacy of Composition [FC]. \normalfont\mdseries This definition identifies all valid instantiations of $(X, \Pi, \Omega)$ that match the rule schema $R$-\texttt{FC}, which captures the fallacy of composition—mistakenly attributing a property of a part to the whole. Let $X = \texttt{chimney}$, $\Pi = \texttt{survives\_fire}$, and $\Omega = \texttt{building}$. Here, $HP(X, \Pi)$ holds: the chimney has the property of surviving fire. In addition, $IPO(X, \Omega)$ holds since the chimney is a structural part of the building, and $LP(\Omega, \Pi)$ holds because the building as a whole lacks the property of surviving fire. The fallacy occurs when one concludes that the entire building must also survive fire merely because one of its components does. This invalid inference results from illegitimately projecting a part's property onto the composite structure.

\begin{equation}
\small
\frac{
\begin{aligned}
\tilde{R}_\Phi = pd(X, \Pi, \Omega) \; &:-\; HP(X, \Pi), \\
&\quad\quad\;\, IPO(X, \Omega), \\
&\quad\quad\;\, LP(\Omega, \Pi)
\end{aligned}
}{
\mathcal{H}_{\tilde{F}_{\text{valid}}} \mapstochar\longrightarrow pd(X, \Pi, \Omega)
}
\tag{R-FC}
\end{equation}

\end{mydefinition}

\begin{mydefinition}\normalfont\bfseries
Begging the Question [BQ].\normalfont\mdseries This definition identifies all valid instantiations of $(X, A)$ that match the rule schema $R$-\texttt{BQ}, which captures the fallacy of begging the question—where a claim is supported by reasoning that ultimately presupposes the claim itself. Let $X = \texttt{bible\_true}$, $A = \texttt{bible\_word\_of\_god}$, and $E = \texttt{bible\_says\_god\_exists}$. Here, $CA(X, A)$ holds: the truth of the Bible is claimed based on the argument that it is the word of God. Then, $EMA(A, E)$ holds: the explicit meaning of that argument is that the Bible asserts God’s existence. Finally, $EMRC(E, X)$ holds: the assertion that God exists relies on assuming the Bible is true. This circular structure results in the fallacy—since the conclusion $X$ is embedded in the reasoning for $X$, the argument does not provide independent support but instead presupposes what it aims to prove, thus invalidating its logical force.

\begin{equation}
\small
\frac{
\begin{aligned}
\tilde{R}_\Phi = pd(X, A) \; &:- \; CA(X, A),\\
&\quad EMA(A, E), \\
&\quad EMRC(E, X)
\end{aligned}
}{
\mathcal{H}_{\tilde{F}_{\text{valid}}} \mapstochar\longrightarrow pd(X, A)
}
\tag{R-BQ}
\end{equation}

\end{mydefinition}

\begin{mydefinition}\normalfont\bfseries
Contextomy [CT].\normalfont\mdseries This definition identifies all valid instantiations of $(\Theta, \Gamma)$ that match the rule schema $R$-\texttt{CT}, which captures the fallacy of contextomy—where a statement is taken out of its original context to support an unrelated conclusion. Let

$\Theta = \texttt{time\_is\_money}$, $M = \texttt{time\_is\_valuable\_as\_money}$,

$\Delta = \texttt{time\_is\_literally\_money}$,
$\Phi = \texttt{third\_world\_countries\_have\_less\_money}$, $\Gamma = \texttt{time\_is\_slower\_in\_third\_world\_countries}$. 

Here, $QC(\Theta, M)$ holds: the phrase “time is money” is reasonably interpreted to mean time is valuable. However, $QOC(\Theta, \Delta)$ holds as well, representing a misreading that treats the phrase out of the context (literally in this case). This misinterpretation is then linked via $FROC(\Delta, \Phi)$ to a socioeconomic fact, and finally $IFQOC(\Phi, \Gamma)$ commits the fallacy by concluding that time moves more slowly in poorer countries. The fallacy arises from detaching a figurative expression from its intended context and chaining it to unrelated empirical claims, thereby producing a logically unsound and rhetorically misleading argument.

\begin{equation}
\small
\frac{
\tilde{R}_\Phi = pd(\Theta, \Gamma) \; :- \;
\begin{aligned}
& QC(\Theta, M), \quad QOC(\Theta, \Delta), \\
& FROC(\Delta, \Phi), \quad IFQOC(\Phi, \Gamma)
\end{aligned}
}{
\mathcal{H}_{\tilde{F}_{\text{valid}}} \mapstochar\longrightarrow pd(\Theta, \Gamma)
}
\tag{R-CT}
\end{equation}
\end{mydefinition}

\begin{mydefinition}\normalfont\bfseries
Inverse Error [IE].\normalfont\mdseries This definition identifies all valid instantiations of $(\Delta, E)$ that match the rule schema $R$-\texttt{IE}, which formalizes the inverse error fallacy—drawing a false implication in the reverse direction of a valid one, particularly when the inverse domain is broader or less constrained. Let $\Delta = \texttt{cycling\_backwards}$, $E = \texttt{gain\_weight}$, $A = \texttt{cycling\_forwards}$, and $B = \texttt{reduce\_weight}$. Here, $CC(A, \Delta)$ and $CC(B, E)$ hold: cycling forwards is the complement of cycling backwards, and reducing weight is the complement of gaining weight. A valid implication exists in the forward direction: $IM(A, B)$—cycling forwards implies weight loss. However, the reverse implication $\neg IM(B, A)$ also holds: losing weight does not imply cycling forwards. If the reverse implication were valid, it would mean that weight loss is exclusively caused by cycling forwards, and that their complements align perfectly. But this is not the case. The core insight of the inverse error rule is that when the domain of $\Delta$ is broader than that of $E$, the complement of $\Delta$ is correspondingly narrower than the complement of $E$—leading to a logical mismatch when inverting the implication.

\begin{equation}
\small
\frac{
\tilde{R}_\Phi = pd(\Delta, E) \; :- \;
\begin{aligned}
& CC(A, \Delta), \quad CC(B, E), \\
& IM(A, B), \quad \neg IM(B, A)
\end{aligned}
}{
\mathcal{H}_{\tilde{F}_{\text{valid}}} \mapstochar\longrightarrow pd(\Delta, E)
} 
\tag{R-IE}
\end{equation}
\end{mydefinition}

\begin{mydefinition}\normalfont\bfseries
Improper Transposition [IT].\normalfont\mdseries This definition identifies all valid instantiations of $(A, B)$ that conform to the rule schema $R$-\texttt{IT}, which captures the improper transposition fallacy—mistakenly reversing an implication by focusing on a shared consequence while ignoring alternative causes. Let $A = \texttt{rainy\_days}$, $B = \texttt{wet\_ground}$, and $X = \texttt{sprinklers\_on}$. In this case, both $IM(A, B)$ and $IM(X, B)$ hold: rainy days and sprinklers each imply wet ground. Crucially, $X \neq A$, and $\neg IM_T(A, X)$, $\neg IM_T(X, A)$ hold—rain and sprinklers are causally independent. The fallacy arises when one incorrectly infers $IM(B, A)$—that wet ground implies rainy days—despite the existence of other sufficient conditions (e.g., sprinklers) that can also cause $B$.

\begin{equation}
\small
\frac{
\begin{aligned}
& \tilde{R}_\Phi = pd(A, B) \; :- \\[-3pt]
& \qquad \begin{array}{@{}l@{}}
    \begin{array}{c}
        \displaystyle \frac{IM(\Xi, \Psi)}{IM_T(\Xi, \Psi)} \quad 
        \displaystyle \frac{IM(\Xi, \mathrm{H}) \quad IM_T(\mathrm{H}, \Psi)}{IM_T(\Xi, \Psi)}
        \quad \text{(R-IMT)} \\[6pt]
    \end{array}  \\
    \rule{0.8\linewidth}{0.4pt} \\[6pt]
    IM(A, B),\quad IM(X, B),\quad X \neq A, \\
    \neg IM_T(A, X),\quad \neg IM_T(X, A)
\end{array}
\end{aligned}
}{
\mathcal{H}_{\tilde{F}_{\text{valid}}} \mapstochar\longrightarrow pd(A, B)
}
\tag{R-IT}
\end{equation}

\end{mydefinition}

\begin{mydefinition}\normalfont\bfseries
Wrong Direction [WD].\normalfont\mdseries This definition identifies all valid instantiations of $(\Pi, X)$ that match the rule schema $R$-\texttt{WD}, which formalizes the wrong direction fallacy—confusing the direction of causality by treating an effect as if it were the cause. Let $X = \texttt{move\_eye\_close\_to\_mirror}$ and $\Pi = \texttt{mirror\_looks\_like\_eye}$. Here, $CS(X, \Pi)$ holds: the visual effect of the mirror resembling an eye occurs solely due to the proximity of the observer’s own eye. There exists no other alternative cause $Z \neq X$ such that $CS(Z, \Pi)$, satisfying the open cause condition $OC(X, \Pi)$. However, $\neg CS(\Pi, X)$ holds: the visual appearance of the mirror does not in turn cause the eye to move closer. The fallacy arises When the unidirectional logic is reversed—asserting that mirrors inherently resemble eyeballs—this misinterprets a self-induced perceptual effect as an intrinsic property of the object being observed.

\begin{equation}
\small
\frac{
\begin{aligned}
& \tilde{R}_\Phi = pd(\Pi, X) \; :- \\[-3pt]
&  \begin{array}{@{}l@{}}
    \begin{array}{c}
        \displaystyle \frac{CS(X, \Pi) \quad \neg (CS(Z, \Pi),\; Z \neq X)}{OC(X, \Pi)}
        \quad \text{(R-OC)} \\[6pt]
    \end{array} \\
    \rule{0.8\linewidth}{0.4pt} \\[6pt]
    OC(X, \Pi), \quad \neg CS(\Pi, X)
\end{array}
\end{aligned}
}{
\mathcal{H}_{\tilde{F}_{\text{valid}}} \mapstochar\longrightarrow pd(\Pi, X)
}
\tag{R-WD}
\end{equation}

\end{mydefinition}

\begin{mydefinition}\normalfont\bfseries
False Cause [FS].\normalfont\mdseries This definition identifies all valid instantiations of $(T, E)$ that match the rule schema $R$-\texttt{FS}, which formalizes the false casue fallacy—wrongly treating the mere temporal or spatial co-occurrence of two events as evidence that one directly produces the other as a substance. Let $T = \texttt{lightbulb\_switch}$, $E = \texttt{darkness\_emission}$, $\Upsilon = \texttt{room\_event}$, and $X = \texttt{absence\_of\_light}$. Here, $HA(\Upsilon, T)$ and $HA(\Upsilon, E)$ hold: both the action of switching the lightbulb and the resulting darkness occur as part of the same observable room event. However, $RC(X, E)$ identifies the real cause of darkness as the absence of light, not the switching action itself. Since $X \neq T$ and $T @< E$ (the switch action precedes the observed effect), the fallacy arises when one concludes that turning off a lightbulb emits darkness as a physical substance. This misrepresents a lack (the absence of illumination) as a generative act, conflating temporal correlation with causal production.

\begin{equation}
\small
\frac{
\tilde{R}_\Phi = pd(T, E) \; :- \;
\begin{aligned}
& HA(\Upsilon, T), \quad HA(\Upsilon, E), \\
& RC(X, E), \quad X \neq T, \quad T @< E
\end{aligned}
}{
\mathcal{H}_{\tilde{F}_{\text{valid}}} \mapstochar\longrightarrow pd(T, E)
} 
\tag{R-FS}
\end{equation}

\end{mydefinition}


\subsubsection{Complete Prompt for \textbf{FallacyGen-Direct} }This section presents a straightforward approach where LLMs are directly prompted to generate sentences that exemplify specific logical fallacies.

\definecolor{lightergray}{gray}{0.95}
\begin{table}[!htb]
\centering
\caption{Collect example sentences for each fallacy type along with their formal definitions}
\label{tab:fallacy_gen}
\begin{tabular}{|>{\columncolor{lightergray}}p{0.9\linewidth}|}
\hline
\textbf{Instruction:} Given the following list of sentences: \\
\{All sentences in the collected dataset that conform to a specific logical fallacy\}

Explain:  
This is $\{fallacy\_tpye\_and\_definition\}$

\textbf{Query:} Task: Generate 20 new sentences similar to the \{fallacy\_type\} fallacy in the given list.\\
\hline
\end{tabular}
\end{table}

\subsubsection{Complete Prompt for Prolog Knowledge Generation}
See Table \ref{tab:prompt_generate_facts}.
\definecolor{lightergray}{gray}{0.95}
\begin{table}[htb]
\small
\centering
\caption{Each fallacy type is paired with its corresponding combination of facts and rules. For example, the \textit{false analogy} fallacy type is associated with specific analogy-related facts and rules that define its logical structure.}
\label{tab:prompt_generate_facts}
\begin{tabular}{|>{\columncolor{lightergray}}p{0.9\linewidth}|}
\hline
\textbf{Instruction:} generate 20 new \{fallacy\_type\} prolog knowledge combinations, below are examples. \\[1em]
\textbf{Query:}\\
\{Prolog\_Facts\}\\
\{Prolog\_rule\}\\
\hline
\end{tabular}
\end{table}

\subsubsection{Complete Prompt for Fallacy Sentence Transformation}
See Table \ref{tab:transformation_prompt_updated}
\begin{table}[!htb]
\small
\centering
\caption{For each fallacy type, the corresponding list of transformed sentences is utilized, and the Prolog query results for the associated fallacy rule are provided as \texttt{prolog\_facts}.}
\label{tab:transformation_prompt_updated}
\begin{tabular}{|>{\columncolor{lightergray}}p{0.9\linewidth}|}
\hline
\textbf{Instruction:} Generate 20 new \{fallacy\_type\} Prolog knowledge combinations. Study the style of the sentences in the provided list and transform the given Prolog facts into natural language sentences that follow a similar style and structure. \\[1em]

\textbf{Query:}\\
List: \\
{[\{list\_of\_sentence\}]}\\[0.5em]
Prolog Facts: \\
\{prolog\_facts\} \\
\hline
\end{tabular}
\end{table}

\subsubsection{Complete Prompt for \textbf{FallacyGen-Prolog} }Extending the previous setup, this section incorporates an additional step in which LLMs are instructed to generate Prolog representations. This serves as a comparative experiment with \tool{}. The prompt is shown in the Table~\ref{tab:fallacy_prolog}.

\definecolor{lightergray}{gray}{0.95}
\begin{table}[htb]
\centering
\caption{The step to generate prolog}
\label{tab:fallacy_prolog}
\begin{tabular}{|>{\columncolor{lightergray}}p{0.9\linewidth}|}
\hline
\textbf{Instruction:} Given the following list of sentences:\\
\{All sentences in the collected dataset that conform to a specific logical fallacy\}
Explain: This is \{fallacy\_tpye\_and\_definition\}\\
\textbf{Query:} \\
Task: Study the \{fallacy\_type\}, generate new Prolog entities that enable generation of more sentences, each sentence must be a  \{fallacy\_type\} fallacy type. 
\\Generate 20 new Prolog knowledge. The 20 new knowledge can be used to make 20 more new, different sentences. Generate prolog queries according to the new knowledge to generate new false premise sentences. \\
Important: \\
1. Include break down the essence of this logical fallacy into your prolog code comments.. \\
2. Group related entities that are used together.
\\ \hline
\end{tabular}
\end{table}

\subsubsection{Evaluated Models Querying Prompt}
\label{app:query}

In this stage, we present detailed definitions of each logical fallacy to the LLMs. A carefully constructed prompt guides the model to assess whether a sentence exhibits any logical fallacy and to identify the most relevant one(s), if applicable. The results are returned in a structured JSON format to support subsequent processing steps. The \textbf{Sentence Quality Evaluator} and \textbf{Sentence Quality Evaluator Validity} tasks are evaluated using a shared prompt. As the prompt is capable of addressing both tasks effectively, we provide a single example to illustrate its design. The prompt is shown in the Table~\ref{tab:query}.

\definecolor{lightergray}{gray}{0.95}
\begin{table}[htb]
\centering
\caption{Prompt for Model to Judge Logical Fallacies}
\label{tab:query}
\begin{tabular}{|>{\columncolor{lightergray}}p{0.9\linewidth}|}
\hline

\textbf{Instruction:}
You're an expert in logic.\\
Here's a categorisation of the 14 logic errors. Given sentences with logical errors, reflect on them and deal with them as required. \\

\{All\_Logical\_Fallacies\_Definitions\}\\
\textbf{Query:} \\
Judge the following element:\\

\{sentence\}

Please return the result in JSON format as follows:

\{
  "sentence": "The input sentence as provided.",
  "logic\_error": "yes or no — indicate whether the sentence contains a logical error.",
  "logic\_fallacies": "List all applicable fallacy categories, ranked by relevance.",
  "details": "Provide a clear and explicit explanation supporting your judgment."
\}

\\ \hline
\end{tabular}
\end{table}

\subsubsection{Sentence Quality Evaluator Scoring Prompt}
\label{app:scoring}
In this section, we provide the prompt specifically crafted for scoring with LLMs. To ensure scoring consistency and reduce randomness, the temperature parameter of the LLMs is fixed at 0 throughout the evaluation process. The prompt is shown in the Table~\ref{tab:socring_falllacy}.

\definecolor{lightergray}{gray}{0.95}
\begin{table}[!htb]
\centering
\caption{Prompt Used by the Model for Scoring}
\label{tab:socring_falllacy}
\begin{tabular}{|>{\columncolor{lightergray}}p{0.9\linewidth}|}
\hline
\textbf{Instruction:} You are a professional logical fallacy evaluator. Your task is to review a file containing sentences that illustrate specific logical fallacies and assign each a score based on how well the sentence demonstrates the intended fallacy (as indicated by its type field). Evaluate them more holistically based on your understanding of how these fallacies manifest in real-world human communication and reasoning. Do not use code-based method. \\
Scoring Guide: \\
- Score 0: The sentence makes no sense or does not exhibit the intended logical error.\\
- Score 1: The sentence shows only minor applicability of the fallacy in its type field. \\
- Score 2: The sentence largely demonstrates the fallacy. \\
- Score 3: The sentence is a perfect example of the logical fallacy as described in its type. \\
Definitions: \\
-\{fallacy\_definitions\}.\\
\textbf{Query:}\\
 Here are 20 sentences.\\
 \{sentences\}

\\ \hline
\end{tabular}
\end{table}

\subsection{Experimentation Appendix}
\subsubsection{Logical Fallacy Detection Ability F1, FPR, FNR Figure}
The F1, FPR, FNR Figure can be found in Figure \ref{fig:fp_fn_f1}.

\begin{figure*}[!t]
    \centering
    \includegraphics[width=1\linewidth]{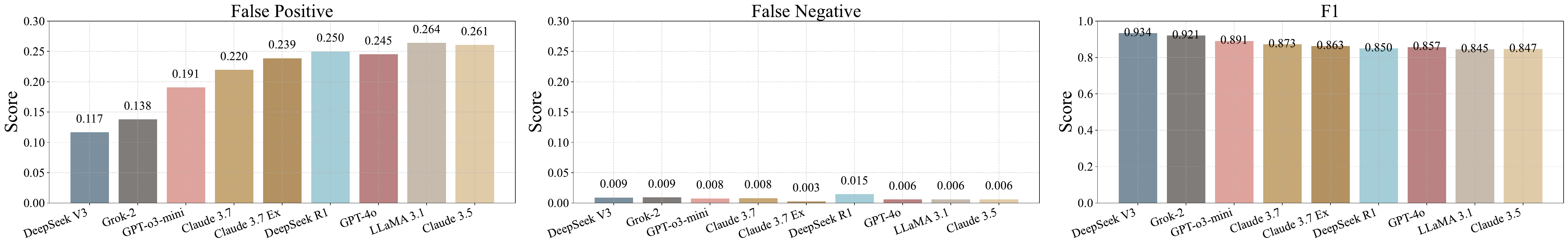}
    \includegraphics[width=1\linewidth]{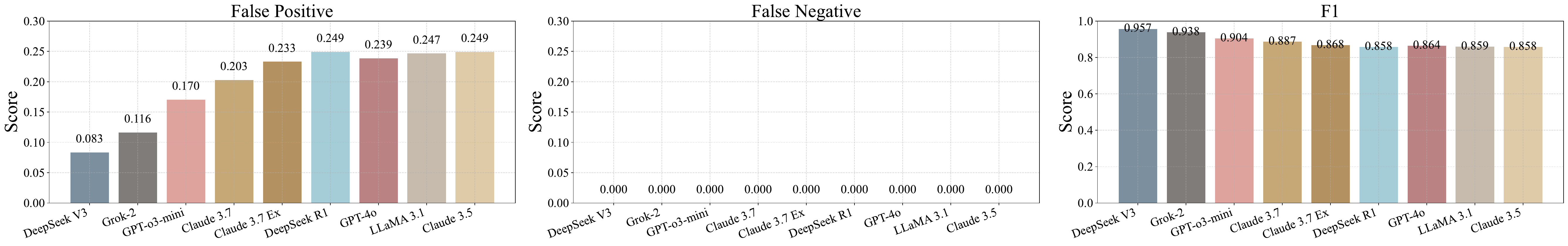}
    \caption{False Positive (lower better), False Negative(lower better), and F1 score(higher better), sorted by F1 score in descending order.  \footnotesize{*Claude 3.7 Ex means Claude 3.7 Extend Thinking. The top section reports the results on \bench, and the bottom section presents those on \augbench. We purposefully use the same y-axis range for FP and FN to show their differences.} } 
    \label{fig:fp_fn_f1}
\end{figure*}

\subsubsection{Sentence Quality Evaluator Validity.}
First, we apply it to all labeled sentences in \bench. As shown in the Table \ref{tab:avg-score}, all fallacy types receive average scores above 2.9, indicating strong discriminative ability. Second, all sentences generated by \tool{} are independently annotated by five human annotators, with high inter-annotator agreement ($Fleiss'\kappa > 0.8$). Following the approach of \citet{ni-etal-2024-afacta}, we compute the average agreement between the LLM and each annotator, yielding Cohen's $\kappa = 0.7699$, which exceeds the commonly accepted threshold for substantial agreement ($\kappa > 0.75$)~\cite{Bujang2017GuidelinesOT, mchugh2012interrater}. These findings support the reliability of our LLM-based evaluator as a valid and scalable proxy for human judgment.

\begin{table}[!ht]
  \centering
  \small
  \caption{Average scores of the three methods on different logic fallacies and their percentage enhancement.\newline
    \footnotesize{*Direct means \textbf{FallacyGen-Direct}, Prolog means \textbf{FallacyGen-Prolog}. Enhance means improvement over sentences generated by Direct using \textbf{\tool{}}.}}
  \label{tab:avg-score}
  \footnotesize
  \setlength{\tabcolsep}{1pt}  

  \begin{tabular}{lrrrr}
    \hline
    \textbf{Fallacies}
      & \textbf{Direct}
      & \textbf{Prolog}
      & \textbf{\tool}
      & \textbf{Enhance (\%)} \\
    \hline
    AC  & 2.32 & 2.48 & \textbf{3.00} & 29.50 \\
    CT  & 2.27 & 2.27 & \textbf{2.92} & 28.68 \\
    IE  & 1.78 & 1.88 & \textbf{2.83} & 58.88 \\
    FP  & 2.60 & 2.63 & \textbf{3.00} & 15.38 \\
    FA  & 2.23 & 2.62 & \textbf{3.00} & 34.33 \\
    WD  & 2.33 & 2.30 & \textbf{3.00} & 28.57 \\
    FC  & 1.97 & 2.42 & \textbf{3.00} & 52.54 \\
    BQ  & 2.10 & 2.45 & \textbf{2.97} & 41.27 \\
    FS  & 2.03 & 2.33 & \textbf{2.95} & 45.08 \\
    IT  & 1.85 & 2.13 & \textbf{3.00} & 62.16 \\
    ID  & 2.27 & 2.28 & \textbf{2.90} & 27.94 \\
    \hline
  \end{tabular}
\end{table}

\subsubsection{Grok-2 Performance Explanation}
This section reports the number of fallacy labels predicted by each model on both \augbench{} and \bench{}. We compute this by summing the number of fallacy types returned per sentence. As shown in the Table \ref{tab:llm_counts}, Grok-2 produces the fewest labels on average, whereas Claude 3.7 extended thinking generates the most, reflecting notable variance in prediction breadth across models.

Interestingly, Grok-2 performs significantly better than expected—it also ranks second in RQ2 for fallacy existence detection. This counter-intuitive result is explained in technical appendix, which shows that Grok-2 tends to generate fewer predicted labels. Producing fewer incorrect labels reduces the overall penalty in the scoring function, which benefits its final categorization score. In contrast, more rigorous models like the Claude 3.7 series generate the highest number of predicted labels for both \bench{} and \augbench{}. However, this also introduces more incorrect labels, negatively impacting their score in the categorization task. This effect is further evidenced in \augbench{}, where the gap in predicted label count between the most verbose model (Claude 3.7 series) and the most conservative (O3-mini) shrinks approximately from 700 to 300 labels .

\begin{table}[!htb]
\centering
\caption{Labels Count comparison for \bench~and \augbench.}
\label{tab:llm_counts}
\resizebox{\columnwidth}{!}{%
\begin{tabular}{lcc}
\toprule
\textbf{LLM} & \textbf{\bench} & \textbf{\augbench} \\
\midrule
Claude 3.7 Ex     & 1836 & 692 \\
Claude 3.7        & 1510 & 578 \\
Llama 3 1.405B    & 1465 & 582 \\
DeepSeek V3       & 1414 & 546 \\
Claude 3.5        & 1416 & 544 \\
DeepSeek R1       & 1405 & 525 \\
GPT-4o            & 1382 & 534 \\
Grok-2            & 1294 & 470 \\
O3 Mini           & 1139 & 414 \\
\bottomrule
\end{tabular}%
}
\end{table}

\subsubsection{Semantic Diversity Analysis}

To assess whether the augmented dataset introduces genuinely novel content rather than paraphrasing the original examples, we conducted a semantic similarity analysis using embeddings from the \texttt{text-embedding-3-large} model~\cite{openai2024embedding}.

We computed cosine similarity scores in the following three settings: \textbf{1. Internal similarity (raw)}: The mean pairwise cosine similarity among the 502 original examples in \textsc{SmartyPat-Bench} is 0.1644. \textbf{2. Internal similarity (augmented)}: The mean pairwise cosine similarity among the 220 generated examples in \textsc{SmartyPat-Bench-Augmented}, constructed via Prolog-based logical rule generation and surface realization, is 0.1936. \textbf{3.Cross-set similarity}: The mean cosine similarity between examples in the original and augmented datasets is 0.1611.

These results demonstrate that both datasets exhibit substantial internal semantic diversity. More importantly, the cross-set similarity remains low, indicating that the generated examples are not simple rephrasings of the original data. In cosine similarity, a score of 1.0 denotes semantic identity, while 0.0 indicates complete divergence. Prior work~\cite{chen2022mcpg,yang2019paws} suggests that paraphrases typically have similarity scores above 0.75 or at least 0.5, both of which are significantly higher than our observed values.

This confirms that the augmented dataset contributes semantically novel structures and does not merely rehash the original content. For full implementation details and reproducibility, please refer to our supplementary materials at: \url{https://anonymous.4open.science/r/supplementary_experiments-5CDB/README.md}.

\begin{table}[h]
    \centering
    \caption{Cosine similarity scores computed using \texttt{text-embedding-3-large}.}
    \label{tab:cosine-similarity}
    \resizebox{\columnwidth}{!}{%
    \begin{tabular}{>{\centering\arraybackslash}p{4.5cm} >{\centering\arraybackslash}p{5cm} >{\centering\arraybackslash}p{4cm}}
        \toprule
        \textbf{Comparison Type} & \textbf{Dataset(s)} & \textbf{Mean Cosine Similarity} \\
        \midrule
        Internal Similarity (Raw)        & \textsc{SmartyPat-Bench}                        & 0.1644 \\
        Internal Similarity (Augmented)  & \textsc{SmartyPat-Bench-Augmented}              & 0.1936 \\
        Cross-set Similarity             & \textsc{SmartyPat-Bench} $\leftrightarrow$ \textsc{SmartyPat-Bench-Augmented} & 0.1611 \\
        \bottomrule
    \end{tabular}%
    }
\end{table}

\section{Limitations and Future Work}

\paragraph{Evaluating LLMs' Capability to Explain Logical Fallacies.}
This work focuses on evaluating LLMs' ability to detect and classify logical fallacies using labeled data. However, assigning a correct fallacy label does not necessarily imply a deep understanding of the reasoning involved. Demonstrating true comprehension would require the model to explain why a sentence constitutes a fallacy—or why it does not. While explanation-based evaluation is an important future direction, it falls outside the scope of this work, which emphasizes automated testcase generation for logical fallacy benchmarking.

\paragraph{Soundness of the Test Oracles.}
Our method relies on LLMs to generate new logical facts, which introduces potential issues with rule conformity. Although the generated facts are often syntactically valid, the underlying \textit{atoms} may not always align precisely with the semantic intent of the \textit{predicate}. As a result, the test oracles—defined as the presence and type of logical fallacies—may not always be perfectly sound. Nonetheless, the generated instances remain effective, and results from RQ1 show that the quality of generated data closely approximates that of manually curated samples.

\end{document}